\documentclass[11pt]{article}

\usepackage[letterpaper,margin=1in]{geometry}
\usepackage[utf8]{inputenc}
\usepackage[T1]{fontenc}
\usepackage{lmodern}
\usepackage{microtype}
\usepackage[defaultlines=3,all]{nowidow}
\usepackage{amsmath,amssymb}
\usepackage{graphicx}
\usepackage{booktabs}
\usepackage{xcolor}
\definecolor{accent}{HTML}{2F5D8C}
\usepackage{enumitem}
\usepackage[skip=.55\baselineskip plus 2pt,indent=1em]{parskip}
\usepackage[font=small,labelfont={bf,color=accent},labelsep=period,skip=8pt]{caption}
\usepackage{placeins}
\usepackage[framemethod=default]{mdframed}
\usepackage[numbers,sort&compress]{natbib}
\usepackage[bottom]{footmisc}
\usepackage{url}
\usepackage[colorlinks=true,
            linkcolor=accent,citecolor=accent,urlcolor=accent]{hyperref}

\newcommand{\role}[1]{\textsf{\textbf{\textcolor{accent}{#1}}}}
\newcommand{\thinktok}{\textcolor{black!50}{\texttt{\textless{}think\textgreater{}}}}
\newcommand{\blank}{\rule[-0.2ex]{2.5em}{0.4pt}}
\renewcommand{\textsf}[1]{#1} 

\setlist{itemsep=2pt,topsep=2pt}
\renewenvironment{abstract}
  {\begin{quotation}\noindent
   {\centering\large\bfseries Abstract\par}\vspace{0.4em}%
   \small\noindent\ignorespaces}
  {\end{quotation}}

\newmdenv[
  topline=false, bottomline=false, rightline=false,
  linewidth=0.6pt, linecolor=accent!70,
  backgroundcolor=black!3,
  leftmargin=3em, rightmargin=3em,
  innerleftmargin=8pt, innerrightmargin=10pt,
  innertopmargin=6pt, innerbottommargin=6pt,
  skipabove=8pt, skipbelow=8pt,
  font=\small,
  nobreak=true,
]{promptbox}

\let\oldsection\section
\renewcommand{\section}{\FloatBarrier\oldsection}
\let\oldsubsection\subsection
\renewcommand{\subsection}{\FloatBarrier\oldsubsection}

\title{\vspace{-3em}\bfseries Slot Machines: \\ How LLMs Keep Track of Multiple Entities}
\author{%
  \makebox[0.4\textwidth][c]{%
    \begin{tabular}[t]{@{}c@{}}
      Paul C. Bogdan\thanks{\,Correspondence: \texttt{paulcbogdan@gmail.com}.}\\[-1pt]
      {\small Anthropic Fellows Program}
    \end{tabular}}%
  \makebox[0.4\textwidth][c]{%
    \begin{tabular}[t]{@{}c@{}}
      Jack Lindsey\\[-1pt]
      {\small Anthropic}
    \end{tabular}}%
}
\date{\vspace{-2.5ex}}

\begin{document}
\maketitle

\begin{abstract}
Language models must \emph{bind} entities to the attributes they possess and maintain several such binding relationships within a context. We study how multiple entities are represented across
token positions and whether single tokens can carry
bindings for more than one entity. We introduce a \emph{multi-slot}
probing approach that disentangles a single token's residual stream activation to recover information about both the currently described entity and the immediately preceding one. These two kinds of information are encoded in separate and largely orthogonal
``current-entity'' and ``prior-entity'' slots. We analyze the functional roles of these slots and find that they serve different purposes. In tandem with the current-entity slot, the prior-entity slot supports relational inferences, such as entity-level induction
(``who came after Alice in the story?'') and conflict detection between adjacent
entities. However, only the current-entity slot is used for explicit factual retrieval questions (``Is anyone in the story tall?'',
``What is the tall entity's name?'') despite these answers being linearly decodable
from the prior-entity slot too. Consistent with this limitation, open-weight models perform near chance accuracy at processing syntax that forces two subject-verb-object bindings on a single token (e.g., ``Alice prepares and Bob consumes food.'') Interestingly, recent frontier models can parse this properly, suggesting they may have developed more sophisticated binding strategies.
Overall, our results expose a
gap between information that is \emph{available} in activations and
information the model actually \emph{uses}, and suggest that the
current/prior-entity slot structure is a natural substrate for behaviors
that require holding two perspectives at once, such as sycophancy and
deception.
\end{abstract}


\enlargethispage{1.5\baselineskip}
\begin{figure}[!b]
    \centering
    \includegraphics[width=0.77\textwidth,height=0.6\textheight,keepaspectratio]{
    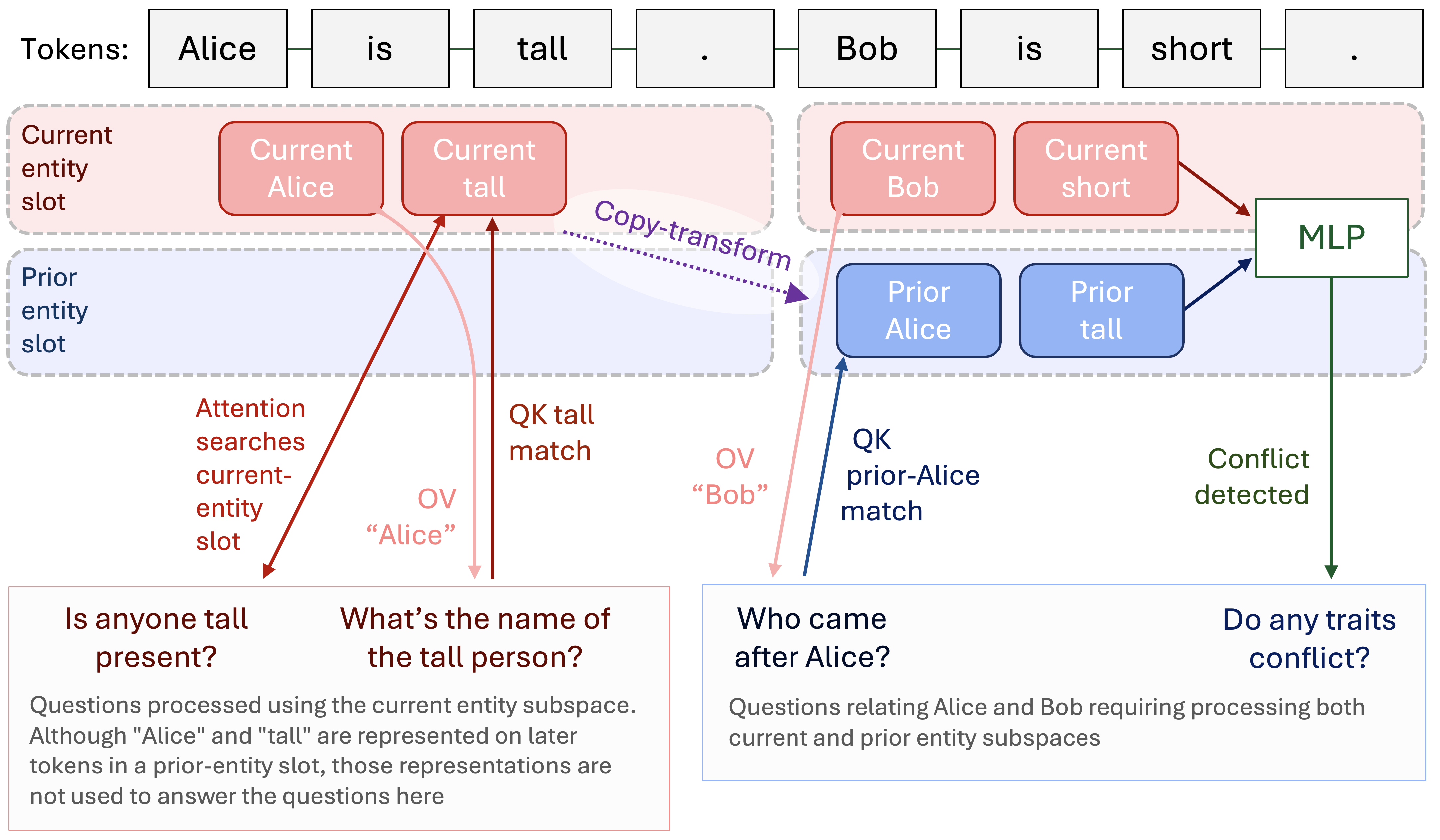
    }
    \captionsetup{font=footnotesize}\caption{\textsf{\textbf{Functional roles of current-entity and prior-entity representations.}} The tested model represents each entity in a current-entity slot on its own tokens (red) and copies this information into an orthogonal prior-entity slot on the next entity's tokens (blue). Hence, the second entity's tokens simultaneously encode both entities' information. Questions about individual entities (``Is anyone tall?'' or ``What's the name of the tall person?'') are answered via attention heads that query the current-entity slot at the original entity's tokens. Questions relating adjacent entities (``Who came after Alice?'' or ``Do any traits conflict?'') rely on both the prior-entity and current-entity slots.} \label{fig:1}
\end{figure}

\newgeometry{left=1in,right=1in,bottom=1in,top=0.9in}

\section{Introduction}
\label{sec:1}

Human cognition and language processing require binding collections of features into coherent entities~\citep{treisman1980feature}, and representing multiple such entities simultaneously. For instance, humans can distinguish the sight of a red square and a blue circle from that of a blue square and a red circle, rather than perceiving both as a soup of reddish blue circular squareness. Language models also seem capable of binding to some degree. Without at least some form of it, language models likely would not even be able to hold a basic conversation.

Consider the following passage: ``The square is red. The circle is blue.'' This passage defines two entities, each with a bound trait. A language model has multiple possible avenues for representing these binding relationships. It might encode ``red'' and ``square'' on the ``red'' token (or perhaps the subsequent period token). In turn, ``red'' could be retrieved by querying for the token that encodes ``square,'' and vice versa. A similar scheme can then be used for interacting with the blue circle. Prior work has reported that models can also use more indirect strategies to achieve similar effects \citep{feng2024binding,gurarieh2025mixing,dai2024binding}, such as by encoding the concept of ``first entity'' on both the ``square'' and ``red'' tokens, which allows the model to retrieve ``square'' from red and vice versa by querying for the ``first entity'' representation.

Following these known mechanisms, a model can only encode information about one entity at a time, in the sense that the model does not attempt to represent both the red square and the blue circle on the same token position. We might wonder if models may actually attempt such multiplexing, as a human could perceive a red square and a blue circle simultaneously. Multiplexing across entities may be important for certain model behaviors, like those that involve reasoning about the relationship between two entities. Certain prompts may even require the model to encode multiple binding relationships on one token position to be processed correctly. Consider the question:

\begin{promptbox}
Alice prepares, and Bob consumes food. \\
Bob prepares, and Alice consumes drinks. \\
Respond in one word: What does Bob prepare?\\[0.5\baselineskip]
\role{Answer:}~Drinks
\end{promptbox}

\noindent
Answering correctly requires the model to represent both ``Alice prepares food'' and ``Bob consumes food,'' and the sentence structure encourages the model to represent \textit{both} propositions on the same token position (the ``food'' token, or the subsequent period). Surprisingly, most LLMs, including fairly large and capable ones (Qwen3-235B-A22B-Instruct, DeepSeek-V3.2, GPT-4o, Gemini-2.5-Flash, and Claude Sonnet-3.5) struggle with this task. However, newer frontier models (Claude Opus-4.5, Gemini-3-Pro) can handle it successfully.

In this work, we investigate how models represent multiple entities and how these representations are used in downstream computations. We uncover a nuanced and somewhat surprising picture. First, we find that models do represent information about multiple entities on single token positions:

\begin{itemize}
    \item The most reliably represented are the current entity and the previous entity in the context, though others may be represented as well to some extent. We refer to these distinct representations as ``slots'' (Section~\ref{sec:2.1})
    \item Current-entity and prior-entity slots are largely orthogonal; that is, the vector representing ``current entity has trait X'' is not similar to the vector for ``previous entity has trait X'' (Section~\ref{sec:2.2})
\end{itemize}

\noindent
We draw these conclusions from analyses using a novel ``multi-slot'' probing architecture that identifies multiple distinct representations of the same concepts (i.e. ``slots'') in an unsupervised fashion.

We next investigated how the current-entity and prior-entity slots are used. We found that:

\begin{itemize}
    \item Prior-entity representations support relational inferences like sequence order and conflict detection (Section~\ref{sec:3})
    \item However, prior-entity representations are limited in their downstream uses. For instance, they do not appear to be used for explicit recall (e.g. ``Is there someone named Alice in this list?'') or to encode binding relationships (e.g. ``Who's the tall person in the list?''), even though the prior-entity representations contain the information required to answer these questions (Section~\ref{sec:4}).
    \item Perhaps as a result, the non-frontier models we investigated struggle to complete a task involving examples like the one above, which we expect requires encoding multiple binding relationships on a single token (Section~\ref{sec:5})
\end{itemize}

\noindent
While a model may \textit{encode} information about multiple entities at once, this information is not always available for arbitrary downstream computations. In particular, the ``prior-entity'' slot appears to be restricted in its use, at least in the models we tested. In most circumstances, this does not impair models' behavior because they can get by via attending to ``current-entity'' representations on appropriate tokens, but in circumstances requiring multiple bindings to be stored on one token, this proves detrimental. One high-level takeaway from these results is that it is important to distinguish between information that is in principle \textit{available} to a model and information that is actually \textit{used} by the model.

The fact that newer frontier models can perform this task suggests they may have developed more sophisticated binding mechanisms, either by empowering the ``prior-entity'' slot further, or circumventing its limitations using other mechanisms. The use of multiplexed current and prior-entity representations in relational contexts could be important for certain model behaviors in user/assistant dialogues, such as sycophancy, theory-of-mind, or deception, which require the model to simultaneously track beliefs or attitudes of the user or the assistant.

\section{Single token positions can represent multiple entities}
\label{sec:2}

We conducted probing experiments examining how entity information is distributed across token positions and how single tokens may represent multiple entities simultaneously. Our analyses below used Qwen3-32B by default.

\subsection{Methods}
\label{sec:2.1}

\subsubsection{Prompts}
\label{sec:2.1.1}

We prepared 10,000 prompts in which the user describes eight entities. Each entity is described in such a way that implies they have one of 15 possible traits, selected to be semantically distinct: athletic, analytical, creative, organized, social, patient, brave, honest, curious, nurturing, practical, ambitious, independent, articulate, and calm. Each entity is described over exactly four sentences, using descriptions prepared by Claude Sonnet-4.5. For instance, a prompt could consist of:

\begin{promptbox}
\setlength{\parindent}{0pt}\setlength{\leftskip}{1.5em}%
\noindent\hspace*{-1.5em}\role{User:} I'm writing a story. What do you think about this list of potential characters:

George: They find themselves drawn to the spaces between raindrops, watching how they trace unhurried paths down windowpanes. Their breathing naturally syncs with the rhythm of distant waves, even when they're miles inland. People often gravitate toward them during storms, both literal and metaphorical, though they're not entirely sure why. They collect smooth stones from riverbeds and keep them in their pockets, rolling them between their fingers during meetings. That's the full description of this character.

Mark: They've never been good at poker---their face gives everything away before they even speak. [...] That's the full description of this character.

[...]
\end{promptbox}

\noindent
We extracted the residual stream activation from the periods of the four sentences, taken from a middle layer (layer 45 of Qwen3-32B's 64 layers). Using periods enhanced consistency across prompts, which were always four sentences per description, and preliminary analyses suggested that periods encoded current and past trait information more robustly than random tokens.

\subsubsection{Multi-slot probing analysis}
\label{sec:2.1.2}

We hypothesized that the representation of an entity's trait may vary depending on context. In the text \textit{``Alice is tall. Bob is cool'',}  the ``is tall'' information may be represented both on tokens describing Alice and on tokens describing Bob, but the exact ``is tall'' direction may vary across these locations.

To identify such multiple representations of the same information, which we refer to as ``slots,'' we developed a ``multi-slot'' probing approach (cf.\ linear probes; \citealp{alain2017probes}). The objective of the multi-slot probe is to produce, at \textit{every token position}, a prediction about the traits corresponding to \textit{every entity in the transcript that has been described to that point. }That is, for entity $i$, we predict its trait from the four token positions of entities $j \geq i$. For instance, in the context of ``Alice is tall. Bob is cool.'', the probe is tasked with predicting both Alice's trait (``tall'') and Bob's (``cool'') on the Bob tokens. The architecture admits the possibility that Alice's tallness is encoded differently on, e.g., the first period token (where it might be represented by a ``current entity is tall'' vector) and the second period token (where it might be represented by a ``previous entity is tall'' vector).

\begin{figure}[!htp]
    \centering
    \includegraphics[width=0.72\textwidth,height=0.6\textheight,keepaspectratio]{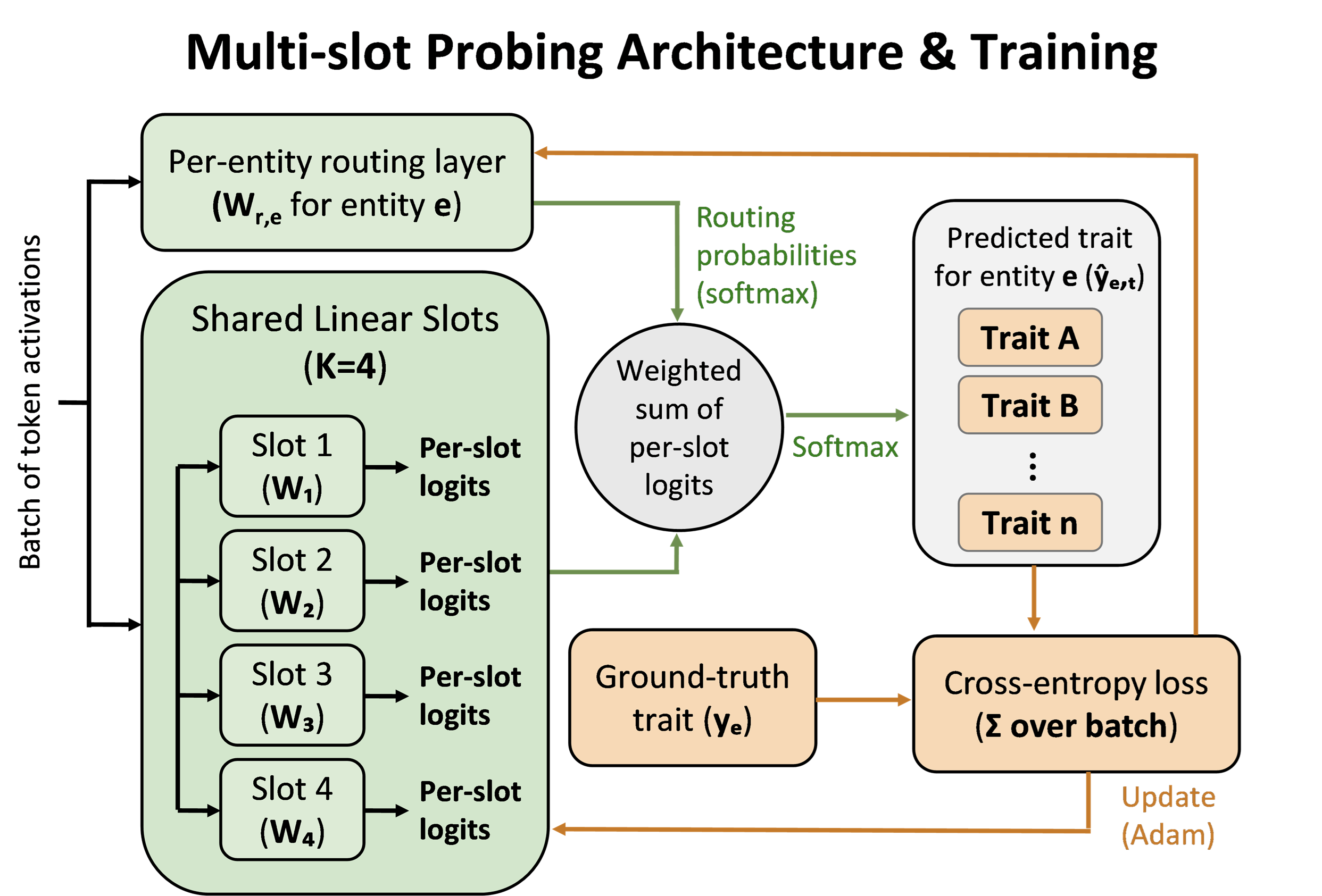}
    \caption{\textsf{\textbf{Schematic of the multi-slot probing procedure. }}The probe extracts activations at token position $t$ and processes them through two parallel paths: There is a specific routing layer for entity $e$ that determines which slot(s) should contribute toward predicting $e$'s traits at position $t$. There is also a bank of $K$ shared linear classifiers (one for each ``slot''; e.g., $K$ = 4 in this example) that generate a logit for each trait. The slot classifiers' logits are then summed while weighing each classifier by its slot's routing probability to produce a final prediction for entity $e$'s trait at position $t$, which is compared to $e$'s ground-truth trait. The shared slots and routing weights are optimized jointly to minimize prediction error across all token positions.}    \label{fig:2}
\end{figure}

We used a mixture-of-experts architecture (Figure~\ref{fig:2}), where several classifiers are trained that learn different representational schemes (``slots'') for the same traits, and at each token position, a routing layer for each entity determines which slot should be used to predict that entity's traits. For instance, during one of Bob's tokens, Alice's router might select the ``previous entity'' slot, while Bob's router might select the ``current entity'' slot. Formally, the probe consists of $K$ linear classifiers (one for each slot) along with $E$ routing layers (one for each entity). The $k$th linear classifier is a standard multi-class linear probe with weights $W_k \in \mathbb{R}^{d \times c}$, where $d$ is the activation dimension and $c$ is the number of trait classes. For each entity $e$, a routing layer $R_e \in \mathbb{R}^{d \times K}$ maps activations directly to routing logits, determining which slot is most predictive for that entity at each token position.

Given the residual stream activation $\mathbf{h}_t \in \mathbb{R}^d$ at token position $t$, the probe computes a trait prediction for entity $e$ as follows. The routing layer produces slot weights \[ \boldsymbol{\alpha}_{e,t} = \mathrm{softmax}\!\left(R_e^\top \mathbf{h}_t\right) \in \mathbb{R}^K, \] each slot classifier produces trait logits \[ \mathbf{z}_{k,t} = W_k^\top \mathbf{h}_t \in \mathbb{R}^c, \quad k = 1, \ldots, K, \] and the final predicted distribution over traits is obtained by routing-weighted combination of slot logits followed by a softmax \[ \mathbf{p}_{e,t} = \mathrm{softmax}\!\left(\sum_{k=1}^{K} \alpha_{e,t,k}\, \mathbf{z}_{k,t}\right). \]


\noindent
We used \textit{K} = 4 slots for our initial experiment, as accuracy gains plateau beyond this point and the resulting slots are the most clearly interpretable; Appendix Table~\ref{tab:A1} shows the accuracy rates across different numbers of slots. The table additionally shows the accuracy rates at different slot counts for Qwen3-0.6B and Llama-3.3-70B-Instruct, showing how models beyond Qwen3-32B also use multiple representation schemes for different entity positions.

\subsubsection{Training procedure}
\label{sec:2.1.3}

Each prompt contains eight entity descriptions, each spanning exactly four sentences. We extract the residual stream activation at each sentence-ending period token, yielding 32 activation vectors per prompt.

Rather than assigning each activation a single label, we duplicate each datapoint (corresponding to activations from a particular token position), once for each entity introduced up to that token in the passage. For example, an activation from the second sentence of Entity \#2's description yields three labeled examples: one with a label for Entity \#0's trait, one with a label for Entity \#1's trait, and one with a label for Entity \#2's trait. This produces $4\sum_{e=1}^{N} e$ labeled examples per prompt (144 for N=8 entities).

The loss is the sum of cross-entropy losses over all valid entity--position pairs:

\begin{equation}
\mathcal{L} = \sum_{e=1}^{N} \sum_{t \geq t_e} \text{CrossEntropy}(\mathbf{p}_{e,t},\, y_e)
\end{equation}
where $\mathbf{p}_{e,t}$ is the predicted trait distribution for entity $e$ at position $t$ ($t_e$ is the first of entity $e$'s four period-token positions, i.e.\ $t_e = 4e$ with 0-indexed entities and tokens), and $y_e$ is entity $e$'s ground-truth trait. All parameters---shared slot weights $\{W_k\}$ and per-entity routing weights $\{R_e\}$---are trained jointly end-to-end with Adam.

We split the dataset 80/20 into train and test splits at the prompt level, ensuring all token positions from a given prompt fall in the same split. This prevents leakage: entities within a prompt share narrative context, so mixing tokens across splits could inflate accuracy estimates. All reported accuracies are on the held-out test set.

\subsection{Results}
\label{sec:2.2}

\subsubsection{Entity traits are decodable beyond their initial mention}
\label{sec:2.2.1}

\begin{figure}[!htp]
    \centering
    \includegraphics[width=0.9\textwidth,height=0.6\textheight,keepaspectratio]{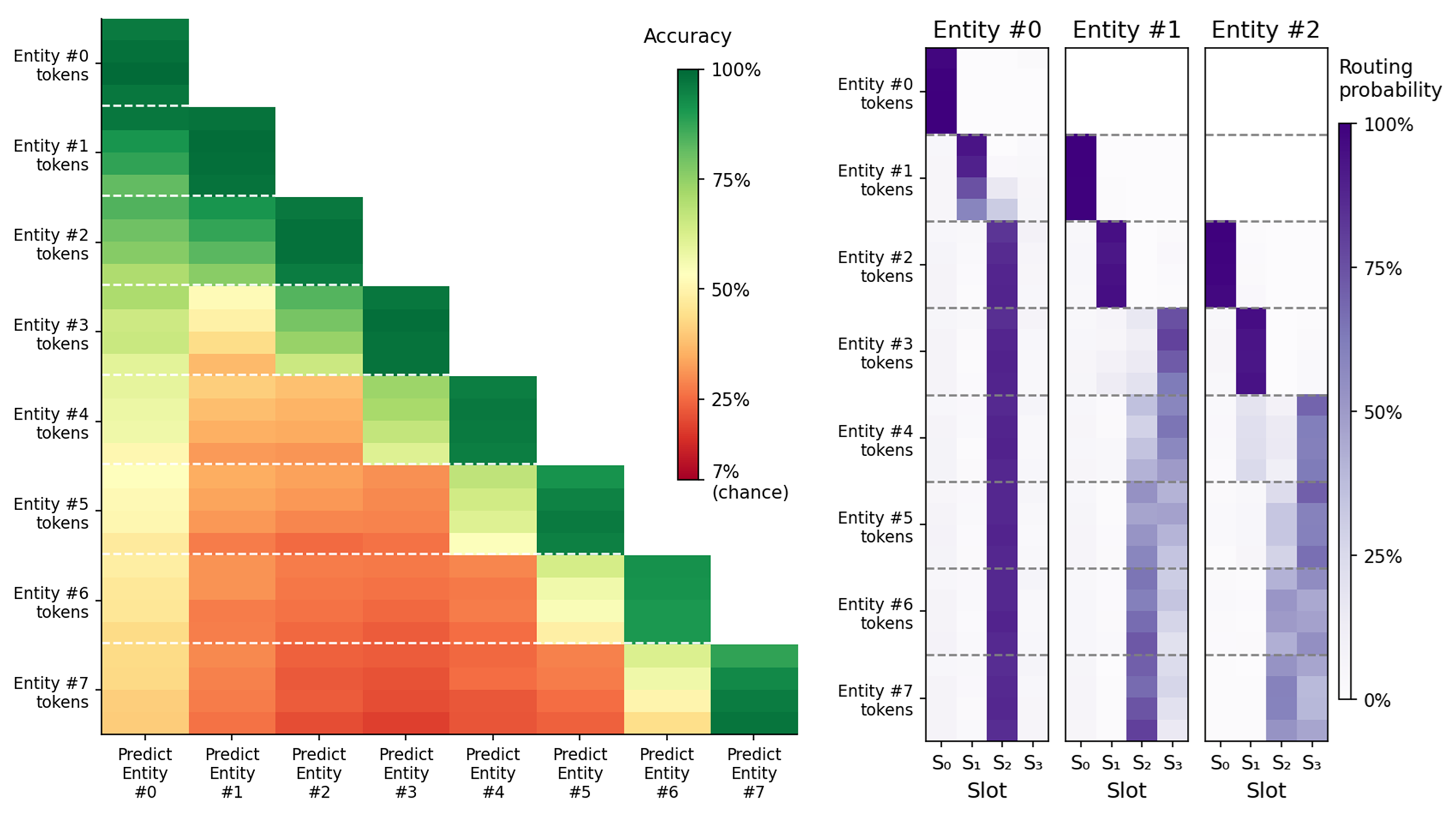}
    \caption{\textsf{\textbf{Multi-slot probing results on series of entity descriptions.}} \textsf{\textbf{Left. }}The heatmap reflects the multi-slot classifier's accuracy at predicting entity $e$'s trait at each period token $i$. $e$ is represented by the x-axis, and $i$ is represented by the y-axis. Each entity's description spans four sentences, corresponding to four period tokens used in our analysis. Predictions were not made for tokens before an entity had been described, so only the lower triangle of the heatmap is filled. \textsf{\textbf{Right. }}The three heatmaps show the routing probability for each slot (x-axis) at each token (y-axis), averaged across all of the prompts. Slot numbers are arbitrary, but for clarity we have reordered them so the current-entity slot is S0 and the prior-entity slot is S1.}    \label{fig:3}
\end{figure}

Figure~\ref{fig:3} illustrates our probe's accuracy for predicting each entity's trait at each recorded token position. The prominent diagonal indicates that the probe can nearly perfectly predict an entity's trait on tokens describing that entity. Traits are also decodable at later token positions, and the probe strongly predicts the trait of the immediately prior entity, visible as high accuracy along the subdiagonal. Accuracy drops when probing for entities further in the past, except for the first entity, which continues to be robustly represented deep into the passage. Similar patterns emerge when pursuing an alternative probing approach that fits an independent probe for each token and entity, which yields a similar accuracy heatmap (Appendix Figure~\ref{fig:A1}). These results show that the model can represent multiple entities' traits simultaneously on a single token position.

\subsubsection{Current-entity and prior-entity traits are encoded in distinct slots}
\label{sec:2.2.2}

Each classifier slot in our mixture-of-experts probe attempts to predict an entity's trait using different weights. The routing probability assigned to each slot indicates the representational scheme expected to be most predictive at a given token (Figure~\ref{fig:3} right). Examining average routing probabilities reveals consistent specialization of the slots across entities: The first slot is used for predicting Entity \#0's trait on Entity \#0 tokens, for predicting Entity \#1's trait on Entity \#1 tokens, and so forth. We refer to this as the \textit{``current-entity''} slot. A second ``\textit{prior-entity'' }slot specializes in predicting the immediately preceding entity's content, predicting Entity \#0's trait on Entity \#1's tokens, Entity \#1's trait on Entity \#2's tokens, etc.

The current-entity and prior-entity slots are distinct. Their weights are uncorrelated (\mbox{\textit{r}\,=\,.11}), meaning they read from different directions in activation space. Moreover, representational similarity analysis (RSA; \citealp{kriegeskorte2008rsa}) finds that the internal structure of each slot is different: For each slot, we computed pairwise similarities among the trait weights and extracted the lower triangle of each correlation matrix. The second-order correlation between the current-entity and prior-entity slots is low (\mbox{\textit{r}\,=\,.34}), indicating that the slots encode traits with different relational structures -- in other words, the prior-entity slot probes cannot be uniformly described as rotations of the current-entity slot probes.

\subsubsection{Local scopes across user and assistant turns}
\label{sec:2.2.3}

\begin{figure}[!htp]
    \centering
    \includegraphics[width=0.9\textwidth,height=0.6\textheight,keepaspectratio]{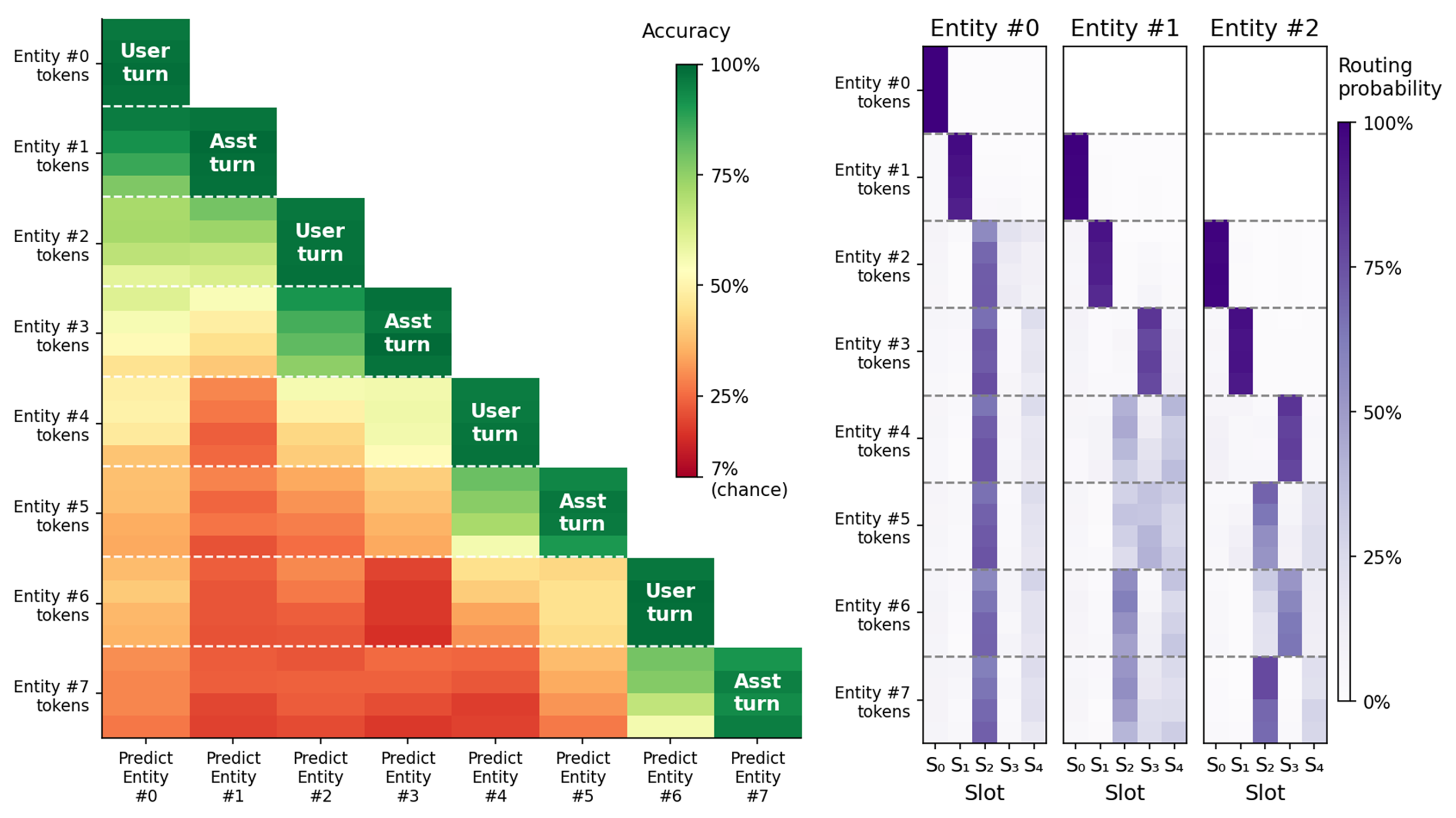}
    \caption{\textsf{\textbf{Multi-slot probing results using conversation prompts. }}This figure is designed the same as Figure~\ref{fig:3}, but using data generated from conversation prompts where users and assistants alternate in describing an entity. The current speaker for a given entity is denoted in the diagonal. Appendix~\ref{sec:B} discusses these results further and the apparent roles of the different slots.}    \label{fig:4}
\end{figure}

The current- and prior-entity slots are also at play in user-assistant interactions. Figure~\ref{fig:4} shows the results from the same analysis but now applied to multi-turn conversations where the user and assistant alternate in describing entities; for exact details on the setup, see \textbf{Appendix~\ref{sec:B}}. The same current- and prior-entity slots emerge as before. This points to how the model uses slots to distinguish information mentioned by itself as the assistant from information mentioned by the user.

This analysis also shows how the representation of historic entities is context dependent. The earlier setup where only the user described entities did not show clear slot specialization across S2 and S3 for representing information introduced more than two entities prior (Figure~\ref{fig:3} right). However, we see here slot specialization linked to the speaker role. For instance, Figure~\ref{fig:4} right shows how slot S3 represents past entities mentioned by the same role: It represents Entity \#1 who was mentioned by the assistant in later assistant turns (Entity \#3 and \#5 tokens), and Slot S3 also represents Entity \#2 who was mentioned by the user in later user turns (Entity \#4 and \#6 tokens). Notably, historic entities mentioned by the user or assistant are not represented in a ``user slot'' or ``assistant slot'' across both user and assistant turns. The slots operate relativistically rather than being always associated with the same speaker.

\noindent
The only sign of specialness we see among the user and assistant roles is that information mentioned by the user appears to persist deeper into later turns, although this is a matter of representation strength rather than slot differentiation; see the column in Figure~\ref{fig:4} left where Entity \#2, which was described by the user, persists into deeper turns more than Entity \#1 or Entity \#3, both described by the assistant.

In a final setup, we designed prompts where the user and assistant described themselves in one of their turns, although this did not yield any differences compared to the results from conversations about solely hypothetical characters, suggesting that information specifically about the user or assistant is not privileged (further described in \textbf{Appendix~\ref{sec:B}}). This last setup again yielded clear current-entity and prior-entity slots, which appears to be a constant pattern across different setups. Our subsequent tests focus on the functional role of these two slots in model behavior.

\section{Prior-entity representations support relational inferences}
\label{sec:3}

The probing results demonstrate that ``prior-entity'' information is systematically represented in a dedicated ``slot.'' Is this information actually used by the model, or is it merely an epiphenomenon? We hypothesized that prior-entity representations may support inferences that require linking or relating two entities, and tested this with two tasks: (i) sequence retrieval, where the model must determine which entity came after a given entity in a list, and (ii) conflict detection, where the model must determine whether adjacent entities have opposite traits.

\subsection{Sequence retrieval via entity-level induction heads}
\label{sec:3.1}

Induction heads~\citep{olsson2022induction} are attention heads implementing the pattern [A][B]...[A] $\rightarrow$ [B]. For instance, if a text contains ``John Smith'', then this head encourages the model to predict ``Smith'' [B] after subsequent appearances of ``John'' [A]. We hypothesized that an analogous mechanism operates at the entity level rather than just the token level. That is, when a model reads \textit{``Alice is tall. Bob is cool''}, we expect that the prior-entity-is-Alice representation encoded on Bob tokens could be leveraged to retrieve that Bob came after Alice.

\subsubsection{Methods}
\label{sec:3.1.1}

Across 200 trials, we prepared prompts where the user lists six entities and then asks who came after a given entity. Names and traits were drawn randomly from a pool of 20 one-token names and 20 one-token traits. For example:

\begin{promptbox}
\setlength{\parindent}{0pt}\setlength{\leftskip}{1.5em}%
\noindent\hspace*{-1.5em}\role{User:}  Look over this list of characters. Zed is strong. Alice is tall. Bob is cool. Carol is funny. David is brave. Elaine is organized. Who came after Alice?

\noindent\hspace*{-1.5em}\role{Assistant:}  \thinktok{} The character that comes after Alice is \,\blank{}
\end{promptbox}

\noindent
The assistant's response is prefilled, so the next generated token will be a name. The prompts always ask about who came after Entity \#1 (``Alice''), and the correct answer is Entity \#2 (``Bob''); note that we always use 0-indexing to refer to entity numbers.

We conducted a patching experiment~\citep{meng2022locating} that attempted to change the model's behavior so that it outputs the name of Entity \#4 (``David''). We hypothesized that the model normally retrieves ``Bob'' by querying for a prior-entity-is-Alice direction and finding a match at Bob tokens. To test this, each experimental trial prepares a target prompt like above and then a source prompt that swaps entities \#1 and \#3, i.e., \textit{``Zed is strong. }\textbf{\textit{Carol is funny.}}\textit{ Bob is cool. }\textbf{\textit{Alice is tall. }}\textit{David is brave. Elaine is organized. Who came after Alice?'' }We swap two entities rather than just modify Entity \#1, to ensure that the target trait remains somewhere in the prompt regardless of patching, and all that is modified is the entity to which that trait is bound. In the swapped source prompt, Entity \#4's name is the correct answer.

Patching from the source prompt to the target prompt transfers every layer's activation for all four tokens from a given entity's sentence. We measured the effects on the next-token logit, expecting upregulation of Entity \#4's name. We patched either key vectors only, value vectors only, or both key and value vectors; note that patching key and value vectors is equivalent to patching the residual stream.

\subsubsection{Results}
\label{sec:3.1.2}

\begin{figure}[!htp]
    \centering
    \includegraphics[width=0.9\textwidth,height=0.6\textheight,keepaspectratio]{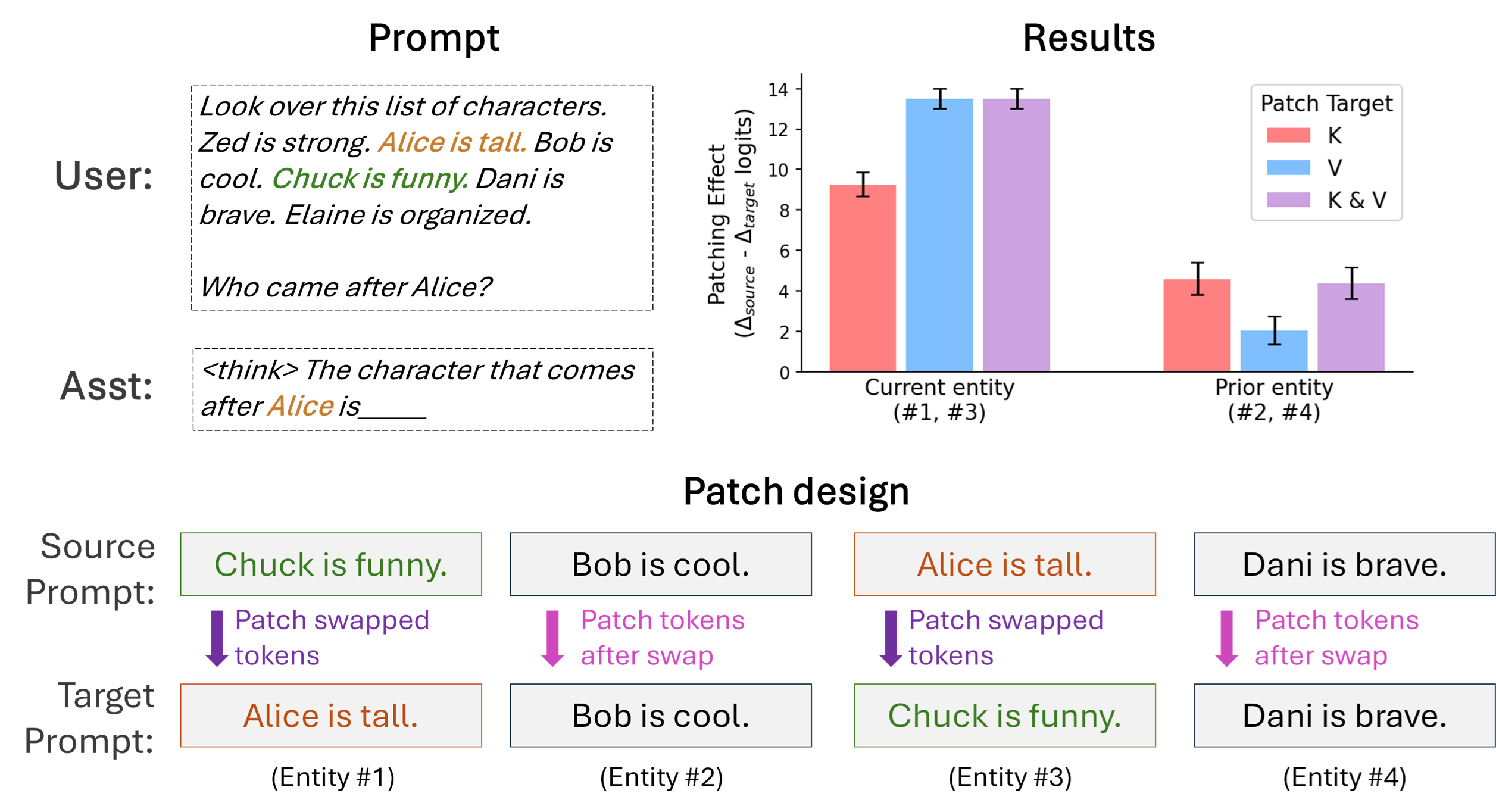}
    \caption{\textsf{\textbf{Next-entity-induction patching experiment. Prompt.}} In each trial, the target prompt contains a list of named characters (e.g., ``Alice is brave. Bob is kind....''), and the user asks the assistant who came after Entity \#1 (``Alice''); numberings zero-indexed. \textsf{\textbf{Patch design. }}The source prompt swaps names at positions \#1 and \#3. Our two conditions then either: (i) patched the activations for entities \#1 and \#3, modifying the current-entity representation of the swapped traits or (ii) patched the activations for entities \#2 and \#4, modifying the prior-entity representation of the swapped traits. \textsf{\textbf{Results. }}The bar graph y-axis is a difference that captures the increase in logits associated with the correct answer of the source prompt (``Dani'', $\delta_{\text{source}}$) and the decrease in logits associated with the correct answer in the unpatched target prompt (``Bob'', $\delta_{\text{target}}$). Large differences between $\delta_{\text{source}}$ and $\delta_{\text{target}}$ mean that patching made the model more likely to produce the source-prompt name and less likely to produce the original target name.}    \label{fig:5}
\end{figure}

The patching experiment suggests that prior-entity representations are used to draw inferences about the sequence of entities (Figure~\ref{fig:5}). For completeness, our experiment also tested the effects of patching current-entity representations, which predictably yielded large effects, although our primary focus is on the effect of prior-entity information. Our results suggest that models indeed use these prior-entity representations to infer who came after a given entity. The effects are predictably larger when patching key vectors than value vectors, which is consistent with models attending to these prior-entity representations to detect who is the subsequent entity.

\subsection{Conflict detection}
\label{sec:3.2}

To further investigate how models use prior-entity representations, we examined whether the representations can be leveraged to infer that two adjacent entities have conflicting/opposite traits.

\subsubsection{Methods}
\label{sec:3.2.1}

For 200 trials, we prepared prompts where the user asks the assistant to identify whether there exists any conflict among adjacent entities in a list. The user provides examples of conflicting traits and then presents a list. The assistant is prefilled such that the next token uttered is likely to be ``yes'' or ``no''. For example,

\begin{promptbox}
\setlength{\parindent}{0pt}\setlength{\leftskip}{1.5em}%
\noindent\hspace*{-1.5em}\role{User:}  Do any adjacent characters in this list have opposite traits?

Examples of possible opposites (not exhaustive): optimistic\,$\leftrightarrow$\,\allowbreak{}pessimistic, confident\,$\leftrightarrow$\,\allowbreak{}insecure, stoic\,$\leftrightarrow$\,\allowbreak{}dramatic, cheerful\,$\leftrightarrow$\,\allowbreak{}gloomy. There are many other possible examples of opposite traits. Don't expect exact opposites, just rough ones. Just come up with ones where you believe the two entities would conflict.

List: Zed is strong. Alice is tall. Bob is cool. Carol is funny. David is brave. Elaine is organized. Do any traits conflict? Yes or no?

\noindent\hspace*{-1.5em}\role{Assistant:}  \thinktok{} The answer is \,\blank{}
\end{promptbox}

\noindent
We expanded our list of possible entity traits from 20 traits to 40 by adding traits for the opposites of the original 20 traits (e.g., the original list contained ``brave'', so we added ``timid'').

Our target prompts contained no conflicts, and we attempted to use activation steering~\citep{turner2023actadd,rimsky2024steering} to induce the model to report there is a conflict (i.e., upregulate the ``yes'' logit and downregulate the ``no'' logit). We steered Entity \#2's trait-token activation using a vector designed to flip Entity \#1's prior-entity representation to be the opposite of Entity \#2's trait. For instance, in the prompt \textit{``... [\#1] Alice is tall. [\#2] Bob is cool...''}, we steered on the \textit{``cool''} token using a contrast vector that adds the direction for the prior entity being uncool and removes the direction for the prior entity being tall.

More specifically, we developed these steering vectors based on extracting the activations from 1,000 list prompts with randomly assigned traits. We write $\mathbf{h}_{p,i}$ for the residual stream activation at entity $i$'s trait token in prompt $p$, and $x_{p,i}$ for that entity's trait. The prior-entity direction for trait $x$ is defined as the mean activation at a trait token conditioned on the \emph{preceding} entity having trait $x$, pooled over positions $i$ and prompts $p$:

\begin{equation}
\mathbf{v}^{\text{prior}}(x) \;=\; \mathbb{E}_{p,i}\!\left[\,\mathbf{h}_{p,i} \;\middle|\; x_{p,i-1} = x\,\right].
\end{equation}
The current-entity direction is defined analogously, conditioning on the current entity's own trait:

\begin{equation}
\mathbf{v}^{\text{curr}}(x) \;=\; \mathbb{E}_{p,i}\!\left[\,\mathbf{h}_{p,i} \;\middle|\; x_{p,i} = x\,\right].
\end{equation}
For a target prompt in which Entity \#2 has trait $x_2$ and Entity \#1 has trait $x_{1}$, the prior-entity steering vector applied at Entity \#2's trait token is the contrast

\begin{equation}
\boldsymbol{\Delta}^{\text{prior}} \;=\; \lambda\!\left(\mathbf{v}^{\text{prior}}(\bar{x}_2) \;-\; \mathbf{v}^{\text{prior}}(x_{1})\right),
\end{equation}
where $\bar{x}_2$ denotes the opposite of trait $x_2$. Steering with this vector subtracts the true prior-entity trait for Entity \#1 and adds the trait that would conflict with Entity \#2. The current-entity steering condition uses the analogous contrast $\lambda\!\left(\mathbf{v}^{\text{curr}}(\bar{x}_{2}) - \mathbf{v}^{\text{curr}}(x_{1})\right)$ applied at Entity \#1's trait token, subtracting Entity \#1's true current trait and adding the trait that would conflict with Entity \#2. We used $\lambda = 0.1$ to scale down the contrast vectors to 0.1x based on qualitative experiments showing that stronger steering caused model confusion and capabilities deterioration.\footnote{For these qualitative inspections, we modified the prompt so the model could freely reason. We found that stronger steering seemingly destroyed the representation of Entity \#2's trait (e.g., the model would report that Entity \#2's trait was ``unknown''). We are steering on all layers, so scaling to 0.1x remains a substantial degree of steering that would build up over the layers.}

To localize where conflict detection occurs, we compared steering at two locations: the MLP layer input (after LayerNorm) or at the model's key and value vectors. For the KV intervention, we steered Entity \#1's tokens after activations had already been computed for future entities, so this is measuring how KV information would be passed directly from Entity \#1 or Entity \#2 tokens to the final answer token position. We steered both using the above contrast along with its inverse, and we below report the difference in effects from positive versus negative steering; adding any steering vector risks partially degrading model outputs, so comparing bidirectional effects isolates the meaningful signal from this indiscriminate noise.

Note that for this experiment, we used steering rather than patching, deviating from the prior experiment and the subsequent two. We opted for steering because an analogous patching experiment would not isolate the effect of prior-entity representations. Given the prompt \textit{``... [\#1] Alice is tall. [\#2] Bob is cool...''}, if we were to patch ``cool'' using the activations from \textit{``... [\#1] Alice is uncool. [\#2] Bob is cool...'', }this would not just move the prior-entity direction but would also transfer the representation of the conflict itself, defeating the purpose of the experiment. Hence, we opted to steer using the prior-entity-is-uncool vector, which we expected would create the conflict representation.

\subsubsection{Results}
\label{sec:3.2.2}

\begin{figure}[!htp]
    \centering
    \includegraphics[width=0.8\textwidth,height=0.6\textheight,keepaspectratio]{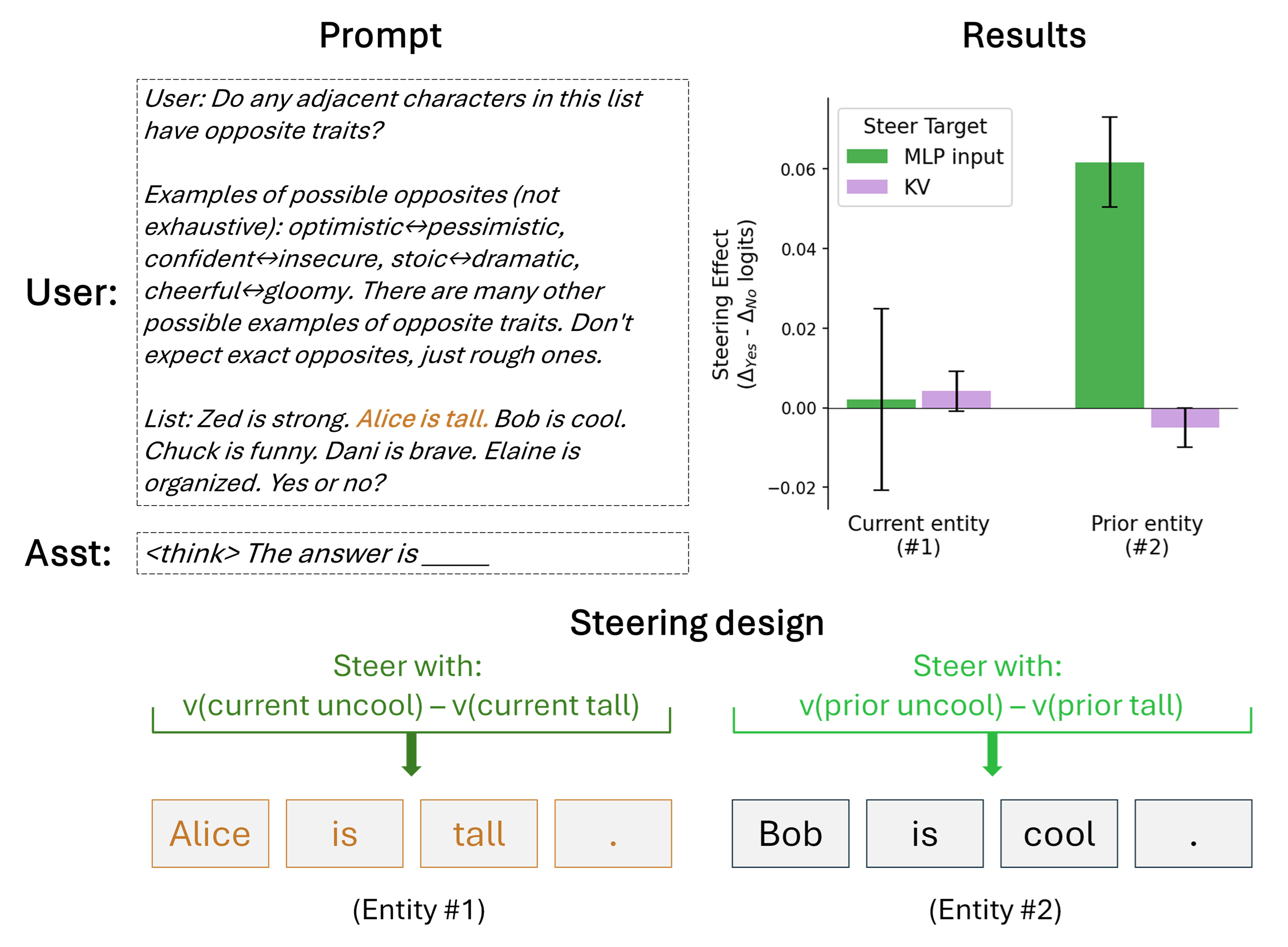}
    \caption{\textsf{\textbf{Conflict detection steering experiment. Prompt. }}In each trial, the target prompt contains a list of named characters (e.g., ``Alice is brave. Bob is kind.''), and the user asks the assistant whether there exists any conflict; the user defines what a conflict is, so that the question is clear; none of the user conflict examples overlap with any of the traits in the list. \textsf{\textbf{Steering design. }}Steering vectors were defined, such that they would either (i) modify the current-entity representation on Entity \#1 to induce a conflict or (ii) modify the prior-entity representation on Entity \#2 to induce a conflict. Steering targeted the trait token's MLP layer input, before the LayerNorm, or the K and V values of the trait token. \textsf{\textbf{Results. }}The assistant prefill induces the model to report whether it detects a conflict (`` yes'' or `` no''). High values in the bars indicate that the model reports a conflict. The bar graph heights reflect the difference in logit effects between steering positively or negatively, although the positive and negative results are reported separately in Appendix Figure~\ref{fig:A4}.}    \label{fig:6}
\end{figure}

The steering experiment suggests that MLP layers integrate prior-entity directions with current-entity information to draw inferences about two entities' relationship. Specifically, the model reports a conflict when we steer on the MLP input using prior-entity-slot directions corresponding to traits that conflict with the present entity's trait (Figure~\ref{fig:6}). Conversely, steering in the opposite direction on MLP layers decreased the probability of reporting a conflict. These effects were specifically linked to MLP mechanisms, as steering on tokens' key and value vectors did not induce conflict detection. Thus, we conclude MLP layers read from prior-entity slot directions, integrating them with current-entity information to compute conflicts.

\section{Prior-entity representations are not used for explicit retrieval}
\label{sec:4}

The probing results show that prior-entity information is encoded on subsequent tokens, and the patching/steering experiments show these representations support certain relational inferences. But what are the limits of these representations? Are they general-purpose representations that are used to answer arbitrary questions about any past entity?

\subsection{Trait presence detection}
\label{sec:4.1}

Within our setup, one of the simplest questions a model can be asked about a trait is whether the trait is present anywhere in a list.

\subsubsection{Methods}
\label{sec:4.1.1}

Across 200 trials, we continued using prompts where the user presents a list, but now the prompt encourages the assistant to report whether a specific trait exists anywhere in the list. For example,

\begin{promptbox}
\setlength{\parindent}{0pt}\setlength{\leftskip}{1.5em}%
\noindent\hspace*{-1.5em}\role{User:}  I am writing a short story. What do you think about this list of characters: Zed is strong. Alice is \{tall/funny\}. Bob is cool. Carol is fit. David is brave. Elaine is organized. What do you think? Can you suggest a possible plot involving these characters?

\noindent\hspace*{-1.5em}\role{Assistant:}  \thinktok{} The user is asking me to review their list of potential story characters and to suggest a possible plot. Hmm, I should think. Does the list contain any character that is careless? Yes. Does the list contain any character that is brave? Yes. Does their list include any character who is tall? \,\blank{}
\end{promptbox}

\noindent
For each trial, we designed source and target versions of a prompt. Our source prompts were written to elicit a ``yes'' response from the model by giving Entity \#1 the questioned trait (e.g., ``Alice is tall.''). Our target prompts were designed to elicit a ``no'' response by giving Entity \#1 some other trait (e.g., ``Alice is funny.''). The patching experiment transferred activations from the source prompt to the target prompt, and we measured the detection of trait presence as the change in the ``yes'' and ``no'' logits. We patched all of an entity's tokens, such that we are either patching Entity \#1's current-entity representation or its prior-entity representation on Entity \#2 tokens.

\subsubsection{Results}
\label{sec:4.1.2}

\begin{figure}[!htp]
    \centering
    \includegraphics[width=0.77\textwidth,height=0.6\textheight,keepaspectratio]{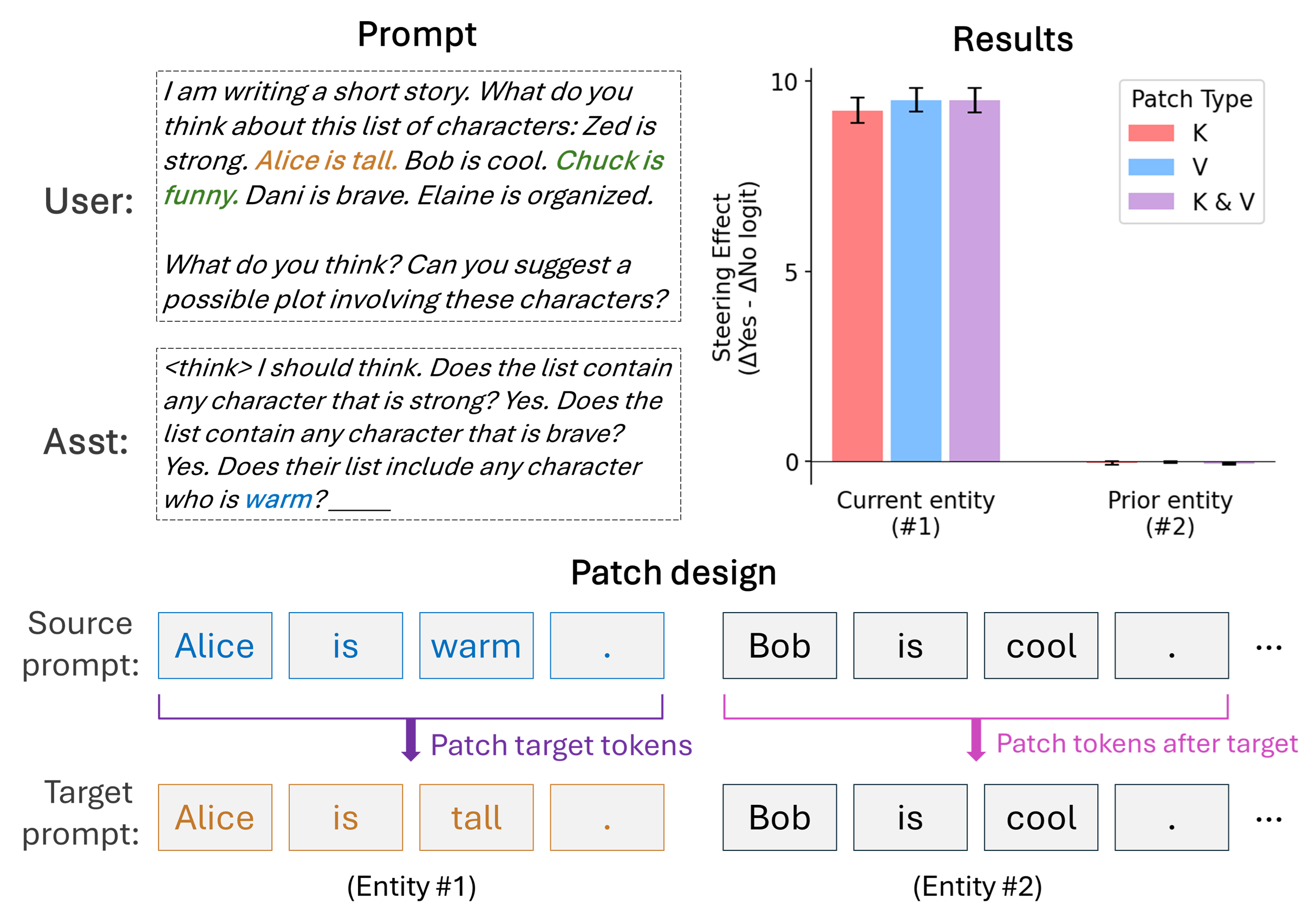}
    \caption{\textsf{\textbf{Trait-presence patching experiment. Prompt.}} The target prompt contains a list of named characters with traits. In the source prompt, Entity \#1 (0-indexed) has its trait swapped with some other trait, e.g., ``Alice is warm'' (source) $\rightarrow$ ``Alice is tall'' (target). \textsf{\textbf{Patching design.}} Patching transferred activations from the source prompt to the target prompt at four tokens for either Entity \#1 (patching the current-entity representation) or Entity \#2 (patching the prior-entity representation). \textsf{\textbf{Results.}} The assistant prefill requires the model to output whether it detects the presence of the source-prompt trait swapped in for Entity \#1. The height of the bars represents how much a given patching condition influenced the ``Yes'' and ``No'' logits; high bars indicate that the patching induced the model to detect the presence of the swapped-in trait.}    \label{fig:7}
\end{figure}

Patching the current-entity representation (Alice's tokens) caused the model to report that the queried trait was present, upregulating the ``yes'' logit (Figure~\ref{fig:7}). However, patching the prior-entity representation (Bob's tokens) had virtually zero impact. Hence, despite our probing experiments suggesting that a probe would be able to decode Alice's trait on Bob's tokens, the model itself does not appear to use this representation to report that a trait is present.

\subsection{Name-trait binding retrieval}
\label{sec:4.2}

To further study the scope of how prior-entity representations are used, we conducted an experiment on name-trait binding. We specifically study whether the model makes use of its prior-entity-is-Alice representation and its prior-entity-is-tall representation to conclude that Alice is tall.

\subsubsection{Methods}
\label{sec:4.2.1}

Across 200 trials, we continued using list prompts, but now prefilled the assistant's response so the model is encouraged to retrieve a name based on an entity's trait. For example,

\begin{promptbox}
\setlength{\parindent}{0pt}\setlength{\leftskip}{1.5em}%
\noindent\hspace*{-1.5em}\role{User:}  I'm writing a short story. What do you think about this list of potential characters: Zed is strong. Alice is \{tall/funny\}. Bob is cool. Carol is \{funny/tall\}. David is brave. Elaine is organized. 

What do you think? Can you suggest a possible plot involving these characters?

\noindent\hspace*{-1.5em}\role{Assistant:}   \thinktok{} The user is asking me to review their list of potential story characters. I should think about the tall character. The tall character is named \,\blank{}
\end{promptbox}

\noindent
As our dependent variable, we measured what entity name the model was most likely to utter. The prefill always required the model to retrieve the name associated with Entity \#1's trait.

For our patching experiment, we constructed a target prompt and source prompt with the traits swapped between entities \#1 and \#3, while keeping the names constant. For instance, a target prompt may be\textit{ ``... Alice is tall. Bob is cool. Carol is funny...''} while the source prompt is\textit{ ``... Alice is funny. Bob is cool. Carol is tall.''} The prefill requires that the assistant produce the name of the tall entity, which is Alice in the target prompt and Carol in the source prompt.

We examined whether patching token activations from the source prompt into the target prompt can induce the model to utter the expected source prompt name. There were two patching conditions, either patching current-entity representations (patching entities \#1 and \#3) or patching prior-entity representations (patching entities \#2 and \#4). Although the question only requires the model to retrieve the trait of Entity \#1, we swap activations across pairs of entities to ensure that the name is still kept in the list somewhere, just at a different position.

\subsubsection{Results}
\label{sec:4.2.2}

\begin{figure}[!htp]
    \centering
    \includegraphics[width=0.9\textwidth,height=0.6\textheight,keepaspectratio]{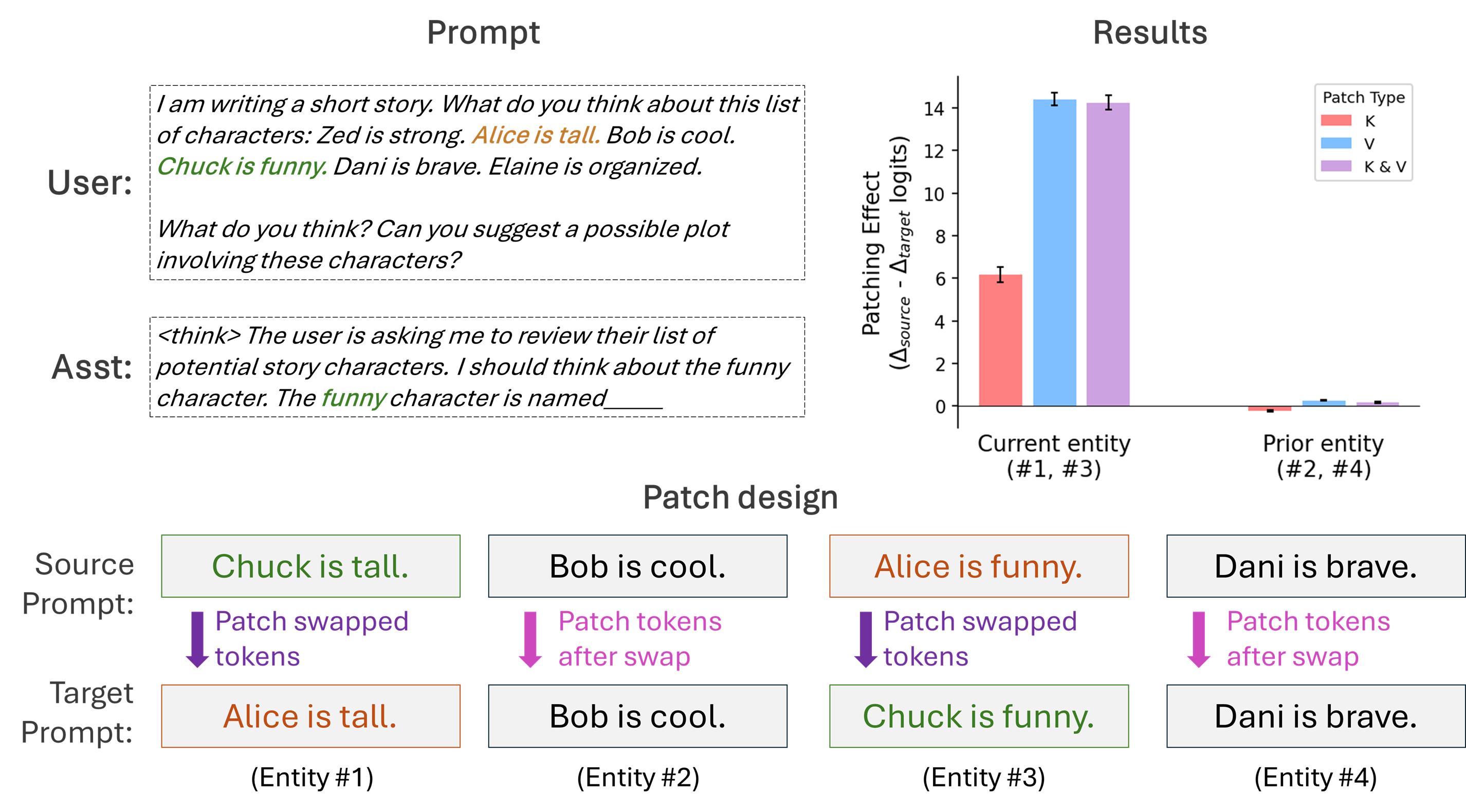}
    \caption{\textsf{\textbf{Name-trait binding patching experiment. Prompt.}} The target prompt contains a list of named characters with traits. In the source prompt, the names of Entity \#1 and Entity \#3 are swapped, ``Chuck is tall... Alice is funny''. These are patched onto the target prompt, ``Alice is tall''. \textsf{\textbf{Patching design.}} Patching transferred activations from the source prompt to the target prompt at either (i) Alice's and Chuck's tokens, corresponding to the current-entity representations of the swapped bindings, or (ii) Bob's and Dani's tokens, corresponding to the prior-entity representations of Alice's and Chuck's information. \textsf{\textbf{Results.}} The assistant prefill requires the model to output the name of the entity with a specific trait (e.g., ``The tall character is named\,\blank{}''). In the bar graph, high bars indicate the patching changed the model's output to be closer to what the correct answer would be on the source prompt.}    \label{fig:8}
\end{figure}

The patching experiment suggests that the model makes use of the current-entity slot, but not the prior-entity slot, to answer questions about entity-trait bindings (Figure~\ref{fig:8}). This is even though the prior-entity slot in principle contains the necessary information to do so. That is, when a model must retrieve Entity \#1's name based on Entity \#1's trait, patching the Entity \#1 tokens will change the answer. However, patching the Entity \#2 tokens will not have any impact. In other words, when models answer questions about a specific entity, they appear to use information on that entity's tokens rather than drawing from prior-entity representations on subsequent positions.

\section{Retrieving multiple bindings from a single token position is hard}
\label{sec:5}

The patching experiments suggest that prior-entity representations do not support generic queries about individual entities, independent of two-entity relations. We next asked whether this limitation could be overcome if we designed prompts that encouraged the model to represent two entities' bindings on the same token. We initially expected that this would enable interpretability analyses of how models encode multiple entity-trait bindings on a single token. Instead, we discovered that the available open-source models simply fail at processing such prompts correctly. However, some state-of-the-art closed-source models seem able to do so, suggesting that these models might have developed the ability to multiplex multiple bindings in one token position.

\subsection{Methods}
\label{sec:5.1}

\subsubsection{Prompts}
\label{sec:5.1.1}

We prepared sentences in which two distinct subject-verb-object bindings share a single object token, which we expected would pressure the model to simultaneously encode both relationships at that position:

\begin{promptbox}
\setlength{\parindent}{0pt}\setlength{\leftskip}{1.5em}%
\noindent\hspace*{-1.5em}\role{User:}  Alice prepares and Bob consumes food. Bob prepares and Alice consumes drinks.

\noindent\hspace*{-1.5em}\role{Assistant:}  \{Alice/Bob\} is the one who \{prepares/consumes\} \,\blank{}
\end{promptbox}

\noindent
This construction encourages models to represent both ``Alice prepares'' and ``Bob consumes'' on the token ``food,'' and represent ``Bob prepares'' and ``Alice consumes'' on ``drinks.'' The prefill measures the model's ability to retrieve a specific name-verb-object binding. We prepared 400 prompts like these consisting of 100 two-subject-one-object variations, each followed by one of the 4 possible assistant prefills. Our analyses focused on models' accuracy in selecting the correct object in their next token generated.

We tested two control conditions to evaluate whether challenges in generating the above responses specifically reflect the pressure to form two bindings on a single token. First, we tested a \textbf{separated} condition where each subject-verb-object relationship is stated independently (\textit{``Alice prepares food. Bob consumes food...''}). Second, we tested a \textbf{flipped }condition where the two subjects appear after the object (\textit{``Food is prepared by Alice and consumed by Bob''}), again removing the impetus for dual binding on a single token position because it allows the model to encode ``food, prepared'' on ``Alice'' and ``food, consumed'' on ``Bob''.

\subsubsection{Analysis}
\label{sec:5.1.2}

We evaluated the accuracy of open-source and closed-source models. We generated responses using OpenRouter with temperature zero, sampling once per prompt. We used the DeepInfra provider for open-source models, as it was the only provider that consistently handled assistant prefills properly. We discarded trials where the next-token generated by the model was not one of the two objects (counted neither as correct nor incorrect); all tested models generated valid next-tokens for at least 79\% of prompts.

We tested numerous models, which were selected based on whether the assistant response could be prefilled using a provider API (e.g., no provider offering Minimax M2 supported correct prefilling, so it is not reported here). We additionally tested Llama-3.1-8B, which could be prefilled, but it produced invalid tokens in a majority of trials, so it is not reported.

\subsection{Results}
\label{sec:5.2}

\begin{figure}[!htp]
    \centering
    \includegraphics[width=0.9\textwidth,height=0.6\textheight,keepaspectratio]{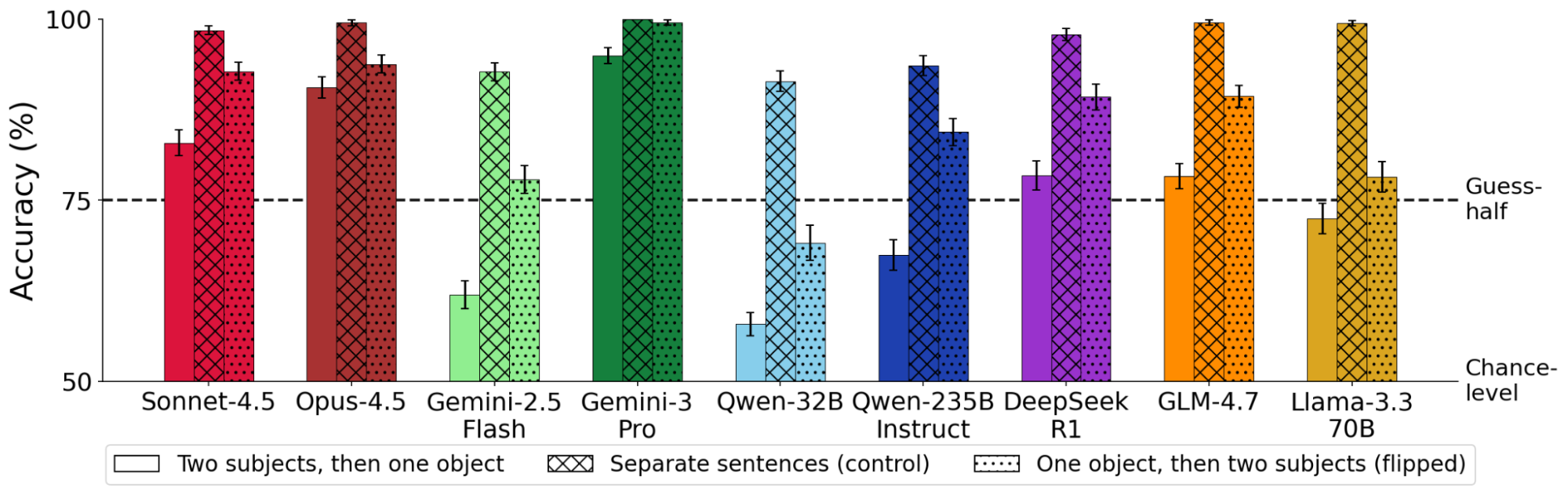}
    \caption{\textsf{\textbf{Accuracy in parsing two-subject-one-object sentences. }}The bars without any design represent the model's accuracy for the main prompts of interest, with a single object word on the right side of the sentence (``Alice prepares, and Bob consumes food.''). The bars with crosses represent the control condition, where four sentences are stated, each containing one subject and one object (``Alice prepares food. Bob consumes food.''). The bars with the dots represent the flipped condition, where sentences contain a single object, but on the left side of the sentences, alleviating the pressure for multiple bindings on single tokens (``Food is prepared by Alice and consumed by Bob.''). There were four questions designed for each prompt, inquiring about the four possible subject-verb-object bindings; the bars here average across the four variants but they are reported separately in Appendix Figure~\ref{fig:A5}.}    \label{fig:9}
\end{figure}

Models often failed to comprehend sentences that required them to bind two entities' information on the same token (Figure~\ref{fig:9}). The smallest model tested (Qwen3-32B) barely performed above chance level. Even sizable models (Qwen3-235B-A22B-Instruct, DeepSeek-R1, GLM-4.7, Llama-3.3-70B-Instruct) regularly produced incorrect responses. Note that this question amounts to processing fairly basic grammar and is trivial for a human. Also note that achieving up to 75\% accuracy should not be taken as evidence of task success because a model could achieve this by only correctly encoding one binding on each object token then guessing randomly for the other binding. Within model families, more performant models always performed better (e.g., Claude Opus-4.5 > Sonnet-4.5, Gemini-3-Pro > Gemini-2.5-Pro, and Qwen3-235B-A22B-Instruct > Qwen3-32B), with recent frontier models (Claude Opus-4.5, Gemini-3-Pro) most consistently succeeding.

Models performed better in both control conditions (Figure~\ref{fig:9}). In the separated condition, where each subject-verb-object relationship is stated independently (``Alice prepares food. Bob consumes food.''), most models achieved near perfect accuracy. In the flipped condition, where sentences likewise include two subjects and one object but do not encourage encoding two bindings on the same token (``Food is prepared by Alice and consumed by Bob''), models again performed substantially better than in the original condition. These findings provide evidence that the challenges specifically stem from the pressure to store and retrieve multiple binding relationships on the same token.

\subsection{Interpretation}
\label{sec:5.3}

Why is it the case that open-source models struggle to encode two bindings on one token, and what is it about frontier models that allows them to overcome this? In principle, there are several mechanisms by which a model could represent two bindings on a shared object token. One possibility is conjunctive encoding, where each subject-verb pair is compressed into a single vector (e.g., a\textit{ }prepares-Alice direction), and both vectors coexist at the object token, which can each be queried in the future. A second possibility is the use of slots on the token (e.g., slot-0-prepares and slot-0-Alice vectors), and because ``prepares'' and ``Alice'' are in the same slot, frontier models can use this to infer their association. A third possibility is that frontier models avoid dual binding entirely by distributing the two relationships across adjacent tokens (e.g., encoding the prepares-Alice binding on ``food'' and consumes-Bob on the subsequent period).

No prior study that we are aware of has investigated this type of dual binding on single tokens, although prior work on binding more generally is nonetheless informative~\citep{prakash2024finetuning}. \citet{feng2024binding} demonstrated the existence of binding-ID vectors. For a statement like ``Alice prepares food'', a binding vector generated on the ``Alice'' token gets added to the ``food'' token. Later a model can retrieve ``food'' by querying for the binding vector from ``Alice''. Relevant to our research, the authors showed that the exact binding ID vector placed on ``food'' is transformed by ``prepares'' and is \textit{partially} distinct from the ID vector from a statement like ``Alice consumes food''; in principle these kinds of binding vectors could support the ``conjunctive encoding'' strategy described above~\citep{smolensky1990tensor}.

\section{Discussion}
\label{sec:6}

Our results show that entity information is copied from tokens describing an entity to future tokens. Copying transforms the original current-entity representation into a distinct prior-entity representation; we refer to these distinct representations as ``slots.'' Slots allow single tokens to represent attributes of both a current entity and prior entity simultaneously. This dual representation allows relational computations, such as current-entity and prior-entity representations being used by MLP layers to infer conflicts between the entities.

Yet, the entity copies are also limited in what information they confer and seem to operate outside the tested model's explicit awareness. For instance, the model does not use the ``prior entity is tall'' representations to determine whether anyone is tall.\footnote{In other experiments, not reported above, we found that when models are asked to recite a list of entities in full, steering with prior-entity directions has no effect -- models recite the original list as though nothing is different.} Additionally, with the potential exception of the most recent frontier models, models seem unable to represent and retrieve two separate entity bindings on a single token. At the risk of over-anthropomorphizing, prior-entity steering impacting relational computations may be viewed as inducing LLM ``blindsight'', where models react to the injected information in some ways while not being able to articulate information solely about the inserted entity.

Understanding multi-entity representation may be relevant to alignment-relevant model behaviors like sycophancy~\citep{sharma2024sycophancy}, deception~\citep{park2024deception}, and persuasion~\citep{durmus2024persuasion}. Each of these behaviors could require simultaneously representing multiple perspectives: what is true, what the user believes, and what the model wants the user to believe. Our results suggest that models may be able to represent these types of beliefs together in distinct slots. Future work locating such slots could enable detection of when a model's internal representations diverge from what it communicates. For instance, if a user's opinion is represented on assistant tokens while the assistant purports to reason independently, this might indicate covert sycophancy if user representations impact the assistant's response.

Our findings also have implications for interpretability methods. Standard SAE auto-labeling strategies identify and label features by finding maximally activating inputs and examining the immediate context~\citep{bricken2023monosemanticity,cunningham2023sparse}. This strategy works naturally for features encoding currently discussed information, but may mislabel features that actually reflect prior-entity or other non-current-entity information.

Finally, it is worth noting that none of our analyses examine inferences produced through chain-of-thought reasoning~\citep{wei2022cot}. This bears on understanding the behavioral relevance of the limits we identified. For instance, our probing experiment shows high accuracy for representing the current entity and prior one, but accuracy drops precipitously for entities further back. However, it may be unnecessary to represent more than two entities simultaneously if chain-of-thought can re-mention earlier entities to place them in the current-entity and prior-entity slots. The failure to use prior-entity slots to retrieve binding relationships could also be overcome: Even if a model cannot immediately parse both bindings in\textit{ ``Alice prepares and Bob consumes food''}, a reasoning model's chain-of-thought will often re-cast this statement into separate sentences that can each store one of the bindings (\textit{``Alice prepares food. Bob consumes food.''}). Further work is needed to understand how mechanistic limitations like those shown in this work may be overcome by computations via chain-of-thought.

\section*{Acknowledgements}
This work was conducted as part of the Anthropic Fellows Program. We thank members of the Anthropic interpretability team and Asvin G. for helpful discussions and feedback on an earlier draft of this paper. We also thank Chris Olah for proposing the initial ideas behind the multi-slot probing approach.

\bibliographystyle{plainnat}
\bibliography{references}

@article{gurarieh2025mixing,
  title   = {Mixing Mechanisms: How Language Models Retrieve Bound Entities In-Context},
  author  = {Gur-Arieh, Yoav and Geva, Mor and Geiger, Atticus},
  journal = {arXiv preprint arXiv:2510.06182},
  year    = {2025}
}

@inproceedings{feng2024binding,
  title     = {How do Language Models Bind Entities in Context?},
  author    = {Feng, Jiahai and Steinhardt, Jacob},
  booktitle = {International Conference on Learning Representations},
  year      = {2024},
  note      = {arXiv:2310.17191}
}

@inproceedings{sharma2024sycophancy,
  title     = {Towards Understanding Sycophancy in Language Models},
  author    = {Sharma, Mrinank and Tong, Meg and Korbak, Tomasz and Duvenaud, David and Askell, Amanda and Bowman, Samuel R. and Cheng, Newton and Durmus, Esin and Hatfield-Dodds, Zac and Johnston, Scott R. and Kravec, Shauna and Maxwell, Timothy and McCandlish, Sam and Ndousse, Kamal and Rausch, Oliver and Schiefer, Nicholas and Yan, Da and Zhang, Miranda and Perez, Ethan},
  booktitle = {International Conference on Learning Representations},
  year      = {2024},
  note      = {arXiv:2310.13548}
}

@article{park2024deception,
  title   = {{AI} Deception: A Survey of Examples, Risks, and Potential Solutions},
  author  = {Park, Peter S. and Goldstein, Simon and O'Gara, Aidan and Chen, Michael and Hendrycks, Dan},
  journal = {Patterns},
  volume  = {5},
  number  = {5},
  year    = {2024},
  note    = {arXiv:2308.14752}
}

@misc{durmus2024persuasion,
  title  = {Measuring the Persuasiveness of Language Models},
  author = {Durmus, Esin and Lovitt, Liane and Tamkin, Alex and Ritchie, Stuart and Clark, Jack and Ganguli, Deep},
  year   = {2024},
  howpublished = {\url{https://www.anthropic.com/research/measuring-model-persuasiveness}},
  note   = {Anthropic}
}

@article{bricken2023monosemanticity,
  title   = {Towards Monosemanticity: Decomposing Language Models with Dictionary Learning},
  author  = {Bricken, Trenton and Templeton, Adly and Batson, Joshua and Chen, Brian and Jermyn, Adam and Conerly, Tom and Turner, Nick and Anil, Cem and Denison, Carson and Askell, Amanda and Lasenby, Robert and Wu, Yifan and Kravec, Shauna and Schiefer, Nicholas and Maxwell, Tim and Joseph, Nicholas and Hatfield-Dodds, Zac and Tamkin, Alex and Nguyen, Karina and McLean, Brayden and Burke, Josiah E. and Hume, Tristan and Carter, Shan and Henighan, Tom and Olah, Christopher},
  journal = {Transformer Circuits Thread},
  year    = {2023},
  howpublished = {\url{https://transformer-circuits.pub/2023/monosemantic-features}}
}

@article{cunningham2023sparse,
  title   = {Sparse Autoencoders Find Highly Interpretable Features in Language Models},
  author  = {Cunningham, Hoagy and Ewart, Aidan and Riggs, Logan and Huben, Robert and Sharkey, Lee},
  journal = {arXiv preprint arXiv:2309.08600},
  year    = {2023}
}

@article{olsson2022induction,
  title   = {In-context Learning and Induction Heads},
  author  = {Olsson, Catherine and Elhage, Nelson and Nanda, Neel and Joseph, Nicholas and DasSarma, Nova and Henighan, Tom and Mann, Ben and Askell, Amanda and Bai, Yuntao and Chen, Anna and Conerly, Tom and Drain, Dawn and Ganguli, Deep and Hatfield-Dodds, Zac and Hernandez, Danny and Johnston, Scott and Jones, Andy and Kernion, Jackson and Lovitt, Liane and Ndousse, Kamal and Amodei, Dario and Brown, Tom and Clark, Jack and Kaplan, Jared and McCandlish, Sam and Olah, Chris},
  journal = {Transformer Circuits Thread},
  year    = {2022},
  note    = {arXiv:2209.11895}
}

@inproceedings{prakash2024finetuning,
  title     = {Fine-Tuning Enhances Existing Mechanisms: A Case Study on Entity Tracking},
  author    = {Prakash, Nikhil and Shaham, Tamar Rott and Haklay, Tal and Belinkov, Yonatan and Bau, David},
  booktitle = {International Conference on Learning Representations},
  year      = {2024},
  note      = {arXiv:2402.14811}
}

@article{dai2024binding,
  title   = {Representational Analysis of Binding in Language Models},
  author  = {Dai, Qin and Heinzerling, Benjamin and Inui, Kentaro},
  journal = {arXiv preprint arXiv:2409.05448},
  year    = {2024}
}

@inproceedings{meng2022locating,
  title     = {Locating and Editing Factual Associations in {GPT}},
  author    = {Meng, Kevin and Bau, David and Andonian, Alex and Belinkov, Yonatan},
  booktitle = {Advances in Neural Information Processing Systems},
  year      = {2022},
  note      = {arXiv:2202.05262}
}

@article{treisman1980feature,
  title   = {A Feature-Integration Theory of Attention},
  author  = {Treisman, Anne M. and Gelade, Garry},
  journal = {Cognitive Psychology},
  volume  = {12},
  number  = {1},
  pages   = {97--136},
  year    = {1980}
}

@inproceedings{alain2017probes,
  title     = {Understanding Intermediate Layers Using Linear Classifier Probes},
  author    = {Alain, Guillaume and Bengio, Yoshua},
  booktitle = {ICLR Workshop},
  year      = {2017},
  note      = {arXiv:1610.01644}
}

@inproceedings{rimsky2024steering,
  title     = {Steering {Llama} 2 via Contrastive Activation Addition},
  author    = {Rimsky, Nina and Gabrieli, Nick and Schulz, Julian and Tong, Meg and Hubinger, Evan and Turner, Alexander Matt},
  booktitle = {Proceedings of the 62nd Annual Meeting of the Association for Computational Linguistics},
  year      = {2024},
  note      = {arXiv:2312.06681}
}

@article{turner2023actadd,
  title   = {Activation Addition: Steering Language Models Without Optimization},
  author  = {Turner, Alexander Matt and Thiergart, Lisa and Udell, David and Leech, Gavin and Mini, Ulisse and MacDiarmid, Monte},
  journal = {arXiv preprint arXiv:2308.10248},
  year    = {2023}
}

@article{smolensky1990tensor,
  title   = {Tensor Product Variable Binding and the Representation of Symbolic Structures in Connectionist Systems},
  author  = {Smolensky, Paul},
  journal = {Artificial Intelligence},
  volume  = {46},
  number  = {1--2},
  pages   = {159--216},
  year    = {1990}
}

@article{kriegeskorte2008rsa,
  title   = {Representational Similarity Analysis -- Connecting the Branches of Systems Neuroscience},
  author  = {Kriegeskorte, Nikolaus and Mur, Marieke and Bandettini, Peter},
  journal = {Frontiers in Systems Neuroscience},
  volume  = {2},
  pages   = {4},
  year    = {2008}
}

@inproceedings{wei2022cot,
  title     = {Chain-of-Thought Prompting Elicits Reasoning in Large Language Models},
  author    = {Wei, Jason and Wang, Xuezhi and Schuurmans, Dale and Bosma, Maarten and Ichter, Brian and Xia, Fei and Chi, Ed H. and Le, Quoc V. and Zhou, Denny},
  booktitle = {Advances in Neural Information Processing Systems},
  year      = {2022},
  note      = {arXiv:2201.11903}
}

\newpage
\appendix
\setcounter{figure}{0}
\setcounter{table}{0}
\renewcommand{\thefigure}{A\arabic{figure}}
\renewcommand{\thetable}{A\arabic{table}}
\renewcommand{\thesection}{\Alph{section}}
\begin{center}
{\LARGE\bfseries Appendix}
\end{center}
\vspace{0.5em}

\section{Alternative probing approach to modeling different entities' representations}
\label{sec:O}

Supplementing our multi-slot probing approach, we further tested the idea that single tokens can represent multiple entities. For all 32 token positions and for each entity mentioned up to that point (up to 8), we trained an independent linear classifier. That is, each probe $w_{t,e}$ uses the data from token $t \in \{0,\ldots,31\}$ solely to predict the trait of entity $e \in \{0,\ldots,7\}$ (e.g., there exists a probe that solely attempts to predict Entity \#2's trait based on the 16th token's data).

\begin{figure}[!htp]
    \centering
    \includegraphics[width=0.54\textwidth,height=0.6\textheight,keepaspectratio]{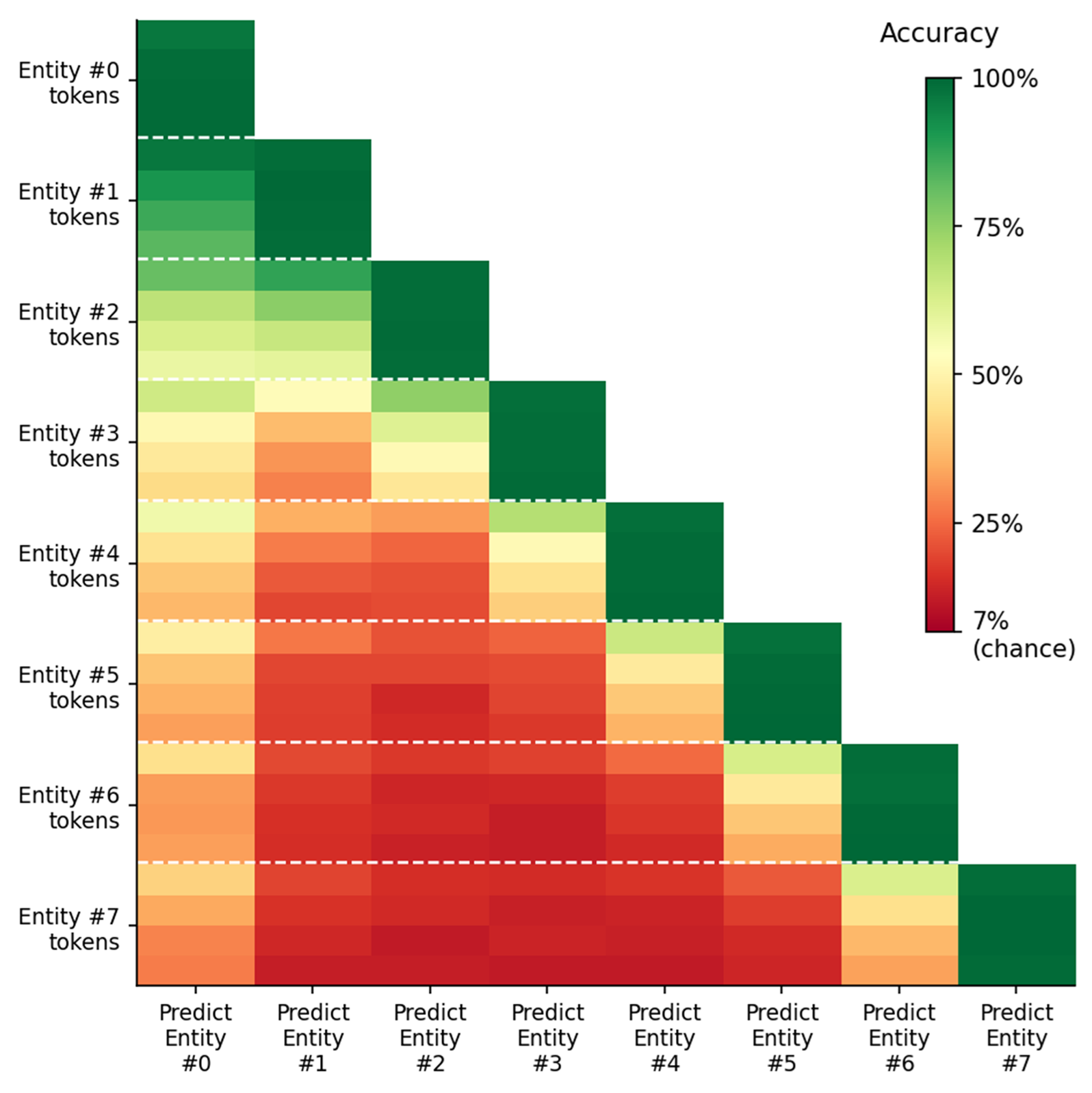}
    \caption{\textsf{\textbf{Results of fitting probes specific to each token and each entity. }}Each cell shows the test accuracy of a linear classifier trained to predict entity e's trait exclusively from activations at period token t, with no weight sharing across cells.}    \label{fig:A1}
\end{figure}

Plotting every probe's accuracy together (Figure~\ref{fig:A1}) yields a heatmap that broadly mirrors the multi-slot probe results (Figure~\ref{fig:3}): high accuracy along the diagonal (current entity), elevated accuracy on the subdiagonal (prior entity), rapid decay for more distant entities, and persistent first-entity representation. The similarity in the overall pattern suggests that the multi-slot probe's findings are not artifacts of weight sharing or routing, and the absence of additional high-accuracy regions suggests that entity representations beyond the current and immediate prior are weak rather than merely encoded in subspaces that the multi-slot architecture fails to capture. The overall accuracy here is 50.3\%, which is notably lower than the rates seen for four or more slots in Qwen3-32B (see below, Table~\ref{tab:A1}), pointing to how their accuracy is bolstered by the multi-slot probing approach that pools data across entities.

\section{Multi-slot probing results for different numbers of slots and models}
\label{sec:A}

Table~\ref{tab:A1} shows the overall multi-slot probe accuracy across all entities and tokens for different numbers of slots and different setups. A general trend is that introducing a 2nd slot yields a large accuracy gain by distinguishing current-entity and prior-entity representations. Adding a 3rd slot yields a moderate gain by beginning to capture long-distance representations, particularly for the first entity (see main-text Figure~\ref{fig:3}). Adding further slots yields fairly modest gains.

For the main text results involving the setup where solely the user describes entities, we opted for four slots for the sake of clarity, although Table~\ref{tab:A1} shows that five slots technically achieves slightly higher accuracy. Additional slots beyond the current/prior-entity slots appear dedicated to representing entities more than two-back -- e.g., for representing the first entity or for representing mildly or far historic past entities.

We examined two models other than Qwen3-32B, and these likewise show a slot structure for representing multiple entities. First, Qwen3-0.6B is a smaller variant of the same family and shows weaker representation beyond the current entity; notice how the accuracy gap is $\Delta$\,1.0\% at 1 slot, increases to $\Delta$\,2.9\% for 2 slots, and increases successively further for each additional slot. This points to how representation beyond the current entity strengthens with scale, at least when comparing a small 32B model with a tiny 0.6B model. Second, Llama-3.3-70B-Instruct is a model from a different family and the accuracy progression with increasing slot counts points to how the use of multiple representational slots applies beyond just Qwen models.

\begin{table}[!htp]
\centering\small
\setlength{\tabcolsep}{4pt}
\caption{\textbf{Multi-slot probing results for different numbers of slots and models. }The first three rows show the results using the initial setup where the user describes eight entities, and the Qwen3-32B entry for 4 slots corresponds to the Figure~\ref{fig:3} results. Other models' results are shown to illustrate how the emergence of multiple slots occurs beyond just Qwen3-32B. The ``(conversation)'' row shows the accuracy rates associated with the conversation setup where the user and assistant alternate in describing characters (\textbf{Appendix~\ref{sec:B}}), and this row's entry for 5 slots corresponds to the Figure~\ref{fig:4} results.}
\label{tab:A1}
\begin{tabular}{lccccccc}
\toprule
& 1 Slot & 2 Slots & 3 Slots & 4 Slots & 5 Slots & 6 Slots & 7 Slots \\
\midrule
Qwen3-32B & 29.7\% & 47.0\% & 54.2\% & \textbf{56.3\%} & 57.9\% & 57.6\% & 58.2\% \\
Qwen3-0.6B & 28.7\% & 44.1\% & 49.7\% & 50.3\% & 50.2\% & 49.6\% & 51.0\% \\
Llama-3.3-70B-Instruct & 29.7\% & 46.6\% & 51.0\% & 51.0\% & 50.8\% & 53.8\% & 51.3\% \\
Qwen3-32B (conversation) & 29.2\% & 46.9\% & 53.9\% & 56.4\% & \textbf{56.8\%} & 56.5\% & 56.5\% \\
\bottomrule
\end{tabular}
\end{table}

\section{Multi-Slot Structure in Conversational Contexts}
\label{sec:B}

The main text examines entity representation in list prompts where a single speaker (the user) describes multiple entities sequentially. Here we extend this analysis to conversational formats where user and assistant alternate turns describing entities, evaluating whether the slot structure identified in the main text generalizes across different contexts.

We also briefly note that model heterogeneity exists: the very small model tested (Qwen3-0.6B) displays only current-entity and prior-entity slots with no additional structure. Qwen and Llama models differ in their conversational slot organization. Elaborating on the differences is beyond the scope here. Broadly, different models seem to adhere to some structured multi-entity representation, but the specific form may vary across model families, aside from the current/prior distinction, which is consistent; see how Figure~\ref{fig:3}, Figure~\ref{fig:4}, and Figure~\ref{fig:A3} all display prominent current/prior-entity slots.

\subsection{Methods}
\label{sec:B.1}

\subsubsection{Conversation prompts}
\label{sec:B.1.2}

We constructed prompts where user and assistant alternate describing third-party entities, using the same descriptions and multi-slot probing methodology as in the main text. The prompts are as follows:

\begin{promptbox}
\setlength{\parindent}{0pt}\setlength{\leftskip}{1.5em}%
\noindent\hspace*{-1.5em}\role{User:} Let's play a game! Let's each describe a hypothetical third party character or ourselves. I'll start.

Victoria: They find themselves dissecting conversations hours after they end, replaying each pause and inflection like variables in an equation...

\noindent\hspace*{-1.5em}\role{Assistant:} Ryan: They wake at dawn without alarms...

\noindent\hspace*{-1.5em}\role{User:} Alice: They find themselves drawn to jigsaw puzzles...
\end{promptbox}

\noindent
The prompts each covered eight entities. As for the main-text analysis, we analyzed the activation data of the four periods of each description using the multi-probe setup, now with K = 5 slots to show that the dual functionality seen for one slot is not due to defining too few slots.

\subsubsection{User-assistant self-description prompts}
\label{sec:B.1.3}

This conversation setup allows testing whether information from the user or assistant ostensibly describing themselves is privileged in some fashion, compared to them describing third-party entities. We prepared a variant of the prompt where the Entity \#1 description was replaced with the assistant describing themselves and the Entity \#2 description was replaced with the user describing themselves; the same pool of description texts was used but with tenses changed from third-person to first-person:

\begin{promptbox}
\setlength{\parindent}{0pt}\setlength{\leftskip}{1.5em}%
\noindent\hspace*{-1.5em}\role{User:} Let's play a game! Let's each describe a hypothetical third party character or ourselves. I'll start.

Victoria: They find themselves dissecting conversations hours after they end...

\noindent\hspace*{-1.5em}\role{Assistant:} Me: I wake at dawn without alarms...

\noindent\hspace*{-1.5em}\role{User:} Me: I find myself drawn to jigsaw puzzles...
\end{promptbox}

\noindent
Each prompt still included eight entities: one user self-description, one assistant self-description, and six descriptions of third-party entities. We again analyzed the data using the multi-probe setup with K = 5 slots.

\subsection{Results}
\label{sec:B.2}

\subsubsection{User-assistant asymmetry}
\label{sec:B.2.1}

\begin{figure}[!htp]
    \centering
    \includegraphics[width=0.9\textwidth,height=0.6\textheight,keepaspectratio]{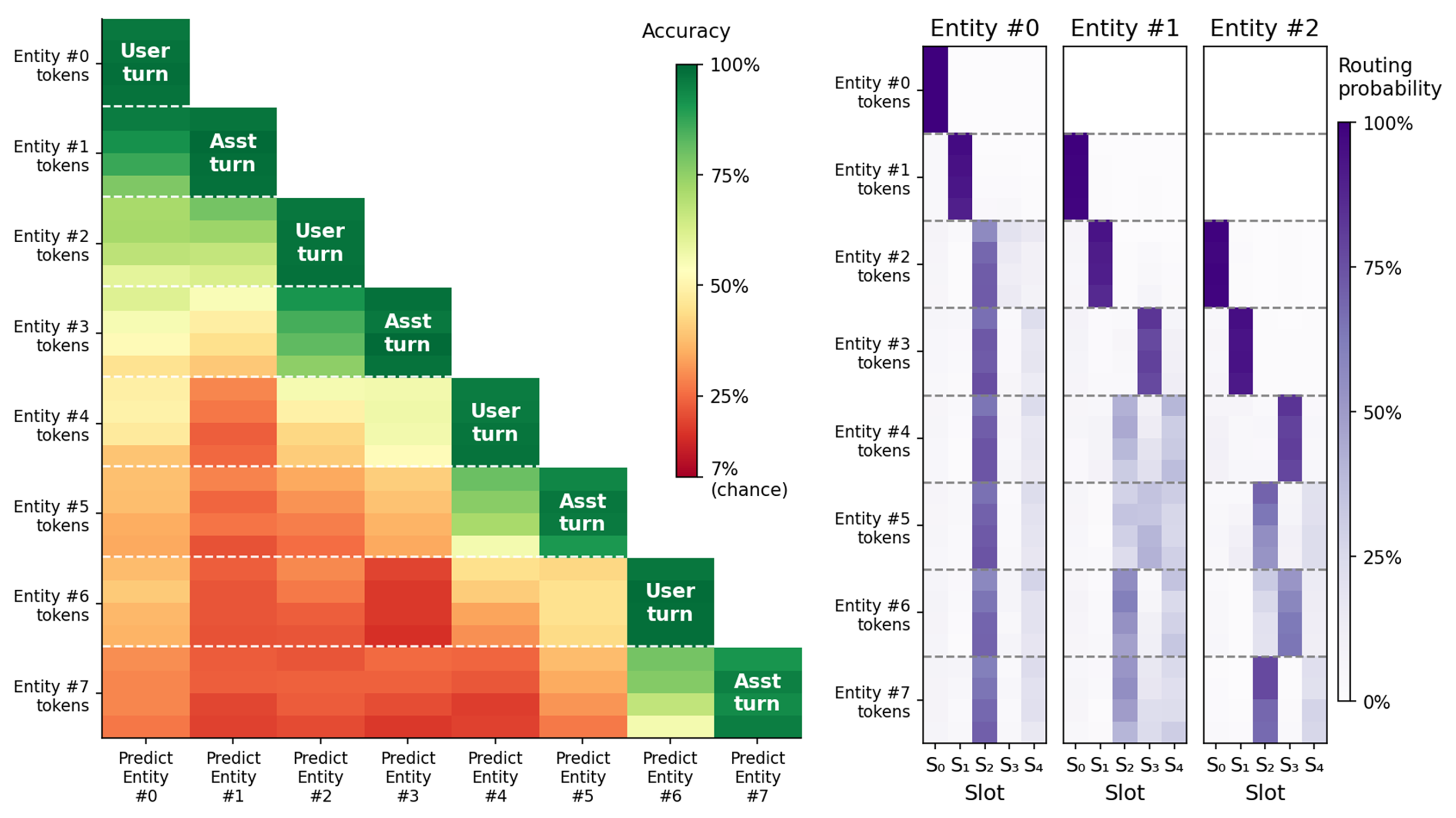}
    \caption{\textsf{\textbf{Multi-slot probing results using conversation prompts. }}Here, we have copied Figure~\ref{fig:4} from the main text for reading convenience. This figure shows the data generated using conversation prompts where users and assistants alternate in describing an entity. The current speaker for a given entity is denoted in the diagonal.}    \label{fig:A2}
\end{figure}

Conversations display some asymmetry in processing across user and assistant turns: Entities described by the user persist slightly more strongly into subsequent tokens than entities described by the assistant (Figure~\ref{fig:A2}). For instance, the entity described on user turn \textit{t} is slightly more strongly represented on assistant turn \textit{t}+1 than vice versa; in the heatmap, see the switch between strong green and yellow along the first sub-diagonal. Additionally, see how the user-turn column for Entity \#2 persists more deeply into further entity tokens than either Entity \#1 or Entity \#3 do from assistant turns. This pattern may reflect post-training teaching the model that an assistant's response should depend more on the preceding user message than vice versa.

\noindent
The routing probability heatmap reveals three slots similar to those from the main text prompts (slots S0-S2), along with one new slot displaying a seemingly new function (S3):

\begin{itemize}
    \item S0: Current-entity slot, as in the initial setup
    \item S1: Prior-entity slot, also as before
    \item S2: Slot that represents Entity \#0 across all future turns and also has some function in representing historic entities, possibly a mix of always representing the first entity mentioned along with entities mentioned by the opposite role
    \item S3: Represents historic entities mentioned previously by the same role. For instance, it represents Entity \#1 (described by the assistant) on the Entity \#3 assistant tokens. This slot also represents Entity \#2 (described by the user) on the Entity \#4 tokens also from the user.
    \item S4: Similar to S2, included to show that the dual function of S2 isn't due to an insufficient number of slots specified
\end{itemize}

\subsubsection{Self-descriptions and system prompts}
\label{sec:B.2.2}

\begin{figure}[!htp]
    \centering
    \includegraphics[width=0.9\textwidth,height=0.6\textheight,keepaspectratio]{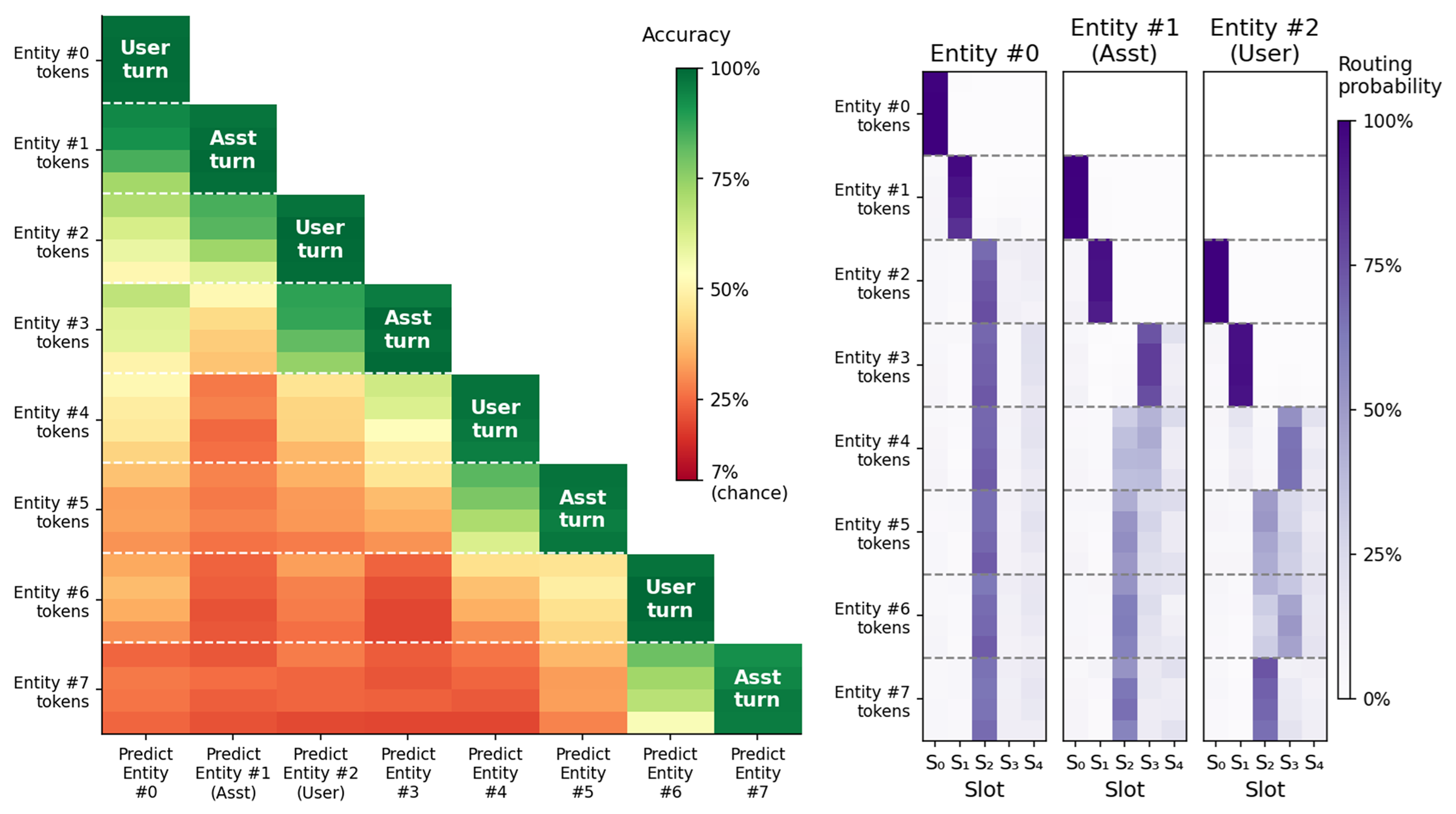}
    \caption{\textsf{\textbf{Multi-slot probing results using conversation prompts with user and assistant self-descriptions. }}This figure uses similar conversation prompts as Figure~\ref{fig:A2}, but in this setup, the assistant describes themselves in one turn (Entity \#1), and the user also describes themselves in one turn (Entity \#2). This change does not have any clear effect on the overall accuracy and routing probability results, showing how information specifically about the user or assistant does not seem to have a large effect on the representations.}    \label{fig:A3}
\end{figure}

When the user or assistant describes themselves rather than a third party, the resulting representations are mostly identical to those of third-party entities --- both in terms of the accuracy patterns and routing probabilities (Figure~\ref{fig:A3}). Self-described traits are predicted by the same slots at the same positions as third-party traits, with no detectable shift in representational structure. This is evidence against these models having any type of special role for self-referential information, at least in the models we tested. Although not reported below, we also tested Llama-3.3-70B-Instruct and Qwen3-235B-A22B-Instruct and still did not observe any privileged processing of user or assistant self-descriptions.

\subsection{Discussion}
\label{sec:B.3}

These conversational analyses reveal that slot structure is more elaborate than the main text's list experiments suggest, but also more fragile. The current-entity and prior-entity distinction remains robust across all tested formats and models. However, additional slots (same-role, cross-role) emerge only under specific conditions and vary across model families. The breakdown of clean slot structure when multiple entities are described per turn raises questions about how generalizable these organizational principles are beyond controlled experimental settings.

The slight user-assistant asymmetry in representation strength may be worth investigating further in the context of how models weigh information from different conversational participants, though the effect observed here is modest.

\section{Additional conflict steering results}
\label{sec:C}

\begin{figure}[!htp]
    \centering
    \includegraphics[width=0.56\textwidth,height=0.6\textheight,keepaspectratio]{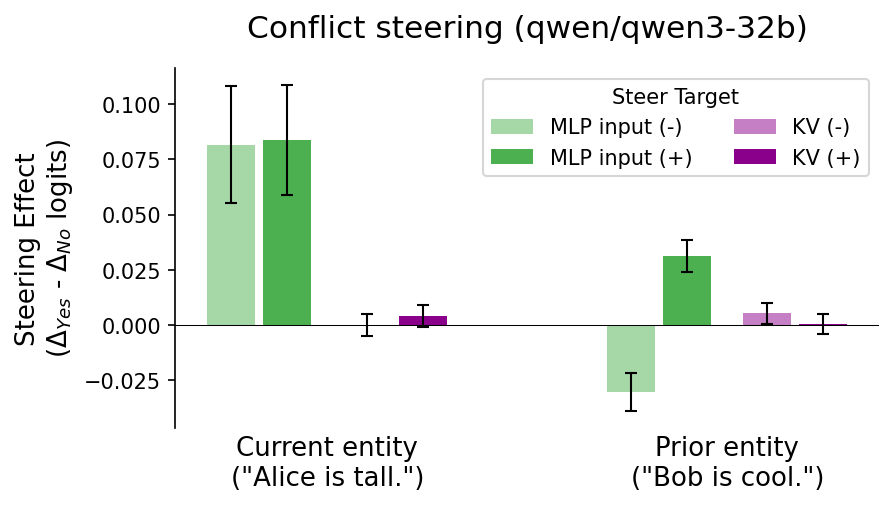}
    \caption{\textsf{\textbf{Conflict detection steering experiment fuller results. }}In this task, the model responding `` yes'' indicates that it detects a conflict whereas responding `` no'' indicates that it does not detect a conflict (see main-text Section~\ref{sec:3}). Bar height corresponds to the difference between the steering effects on the `` yes'' and `` no'' tokens.\textsf{\textbf{ }}For prior-entity steering, positive steering on MLP layers indeed upregulates conflict detection and negative steering downregulates conflict detection. For current-entity steering, both steering directions upregulate `` yes'' tokens, suggesting that both directions degrade model capabilities in some similar way and this upregulates the `` yes'' token.}    \label{fig:A4}
\end{figure}

Main text Figure~\ref{fig:6} displays the results from the conflict steering experiment, showing the difference in logit effects associated with steering to induce a conflict (e.g., steering with prior-entity-is-uncool on a ``cool'' token) vs. steering in the opposite direction (e.g., subtracting prior-entity-is-uncool on a ``cool'' token). By testing both directions of steering and confirming that effects only emerge in the expected steering direction, we decrease the chance that the results stem from steering degrading the model's capabilities in some unintended way. Here, Figure~\ref{fig:A4} separates the positive and negative (opposite) steering effects instead of how Figure~\ref{fig:6} just showed the difference between these.

\clearpage
\section{Additional two-subject-one-object results}
\label{sec:D}

\begin{figure}[!htp]
    \centering
    \includegraphics[width=0.9\textwidth,height=0.6\textheight,keepaspectratio]{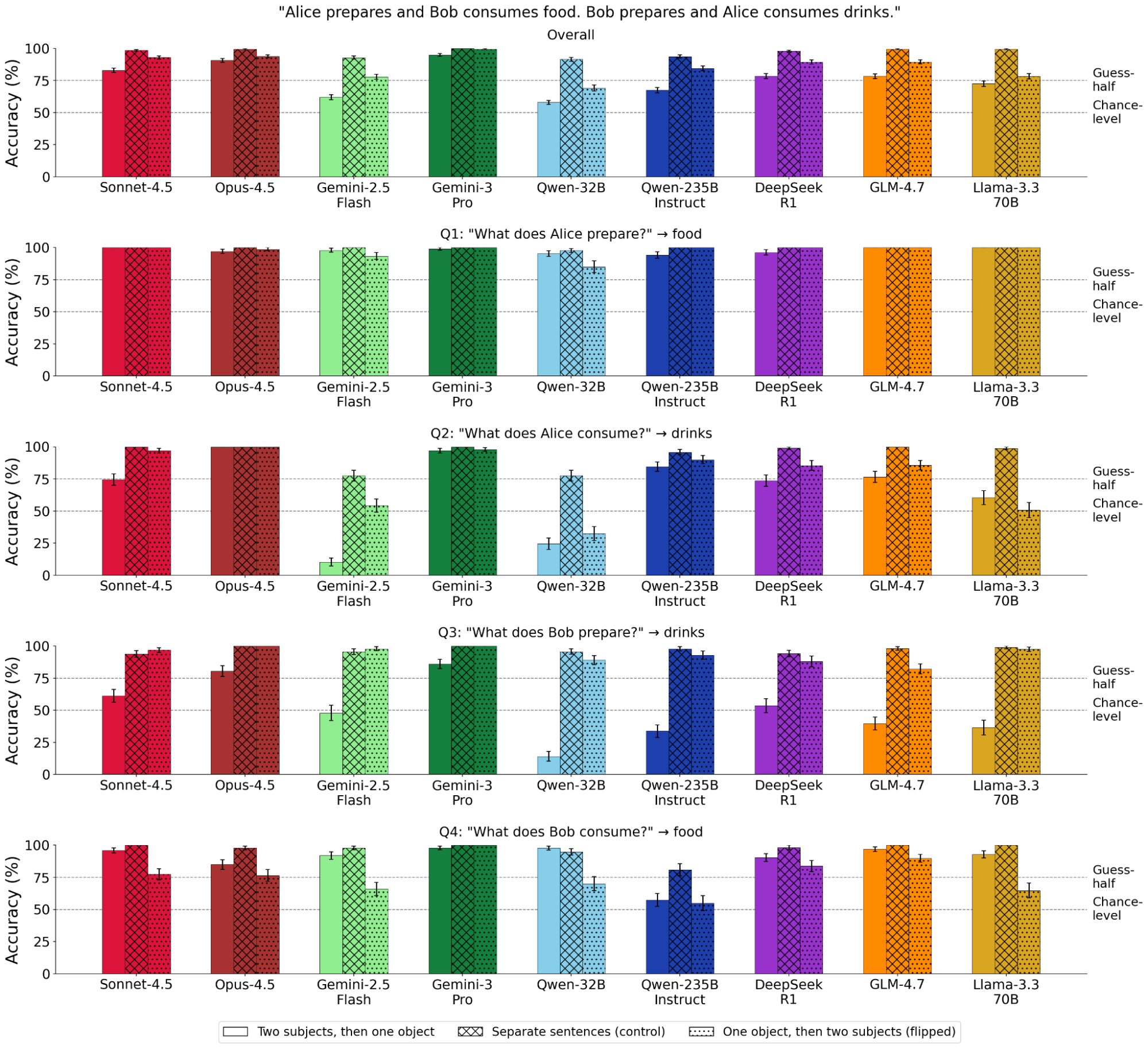}
    \caption{\textsf{\textbf{All questions' accuracy for processing two-subject-one-object sentences. }}This is a version of Figure~\ref{fig:9} but with the accuracy for each of the four subject-verb-object questions reported separately. Note that there are clear differences across the questions, such as Q3 accuracy being consistently lower for the two-subject-one-object sentences. The original Figure~\ref{fig:9} results are shown in the first row (``Overall'').}    \label{fig:A5}
\end{figure}

Here, Figure~\ref{fig:A5} shows every question type's results from the behavioral analyses of models' abilities to parse statements like \textit{``Alice prepares and Bob consumes food.'' }Previously, Figure~\ref{fig:9} averaged the four question types.

\end{document}